%% file: main.tex
\documentclass[10pt,twocolumn,letterpaper]{article}

\usepackage[pagenumbers]{iccv} %

\input{preamble}

\usepackage[accsupp]{axessibility}
\usepackage{multirow}
\usepackage{booktabs}
\usepackage{bbm, dsfont}

\definecolor{iccvblue}{rgb}{0.21,0.49,0.74}
\usepackage[pagebackref,breaklinks,colorlinks,allcolors=iccvblue]{hyperref}

\def\paperID{3473} %
\def\confName{ICCV}
\def\confYear{2025}

\DeclareRobustCommand{\modelname}{\textsc{Semanticist}\xspace}
\DeclareRobustCommand{\armodelname}{$\epsilon$LlamaGen\xspace}

\title{``Principal Components" Enable A New Language of Images}

\author{
Xin Wen$^{1*}$ \authorskip Bingchen Zhao$^{2*\dagger}$ \authorskip Ismail Elezi$^{3}$ \authorskip Jiankang Deng$^{4}$ \authorskip Xiaojuan Qi$^1$$^\ddagger$\vspace{.1em}  \\
{\small $^*$Equal Contribution \authorskip $^\dagger$Project Lead \authorskip $^\ddagger$Corresponding Author}\vspace{.2em}\\
\normalsize
$^1$University of Hong Kong \authorskip 
$^2$University of Edinburgh \authorskip
$^3$Huawei London Research Centre \authorskip
$^4$Imperial College London \\
\small \href{https://visual-gen.github.io/semanticist}{https://visual-gen.github.io/semanticist}
}

\begin{document}
\twocolumn[{%
\renewcommand\twocolumn[1][]{#1}%
\maketitle
\centering
\captionsetup{type=figure}
\includegraphics[width=\linewidth]{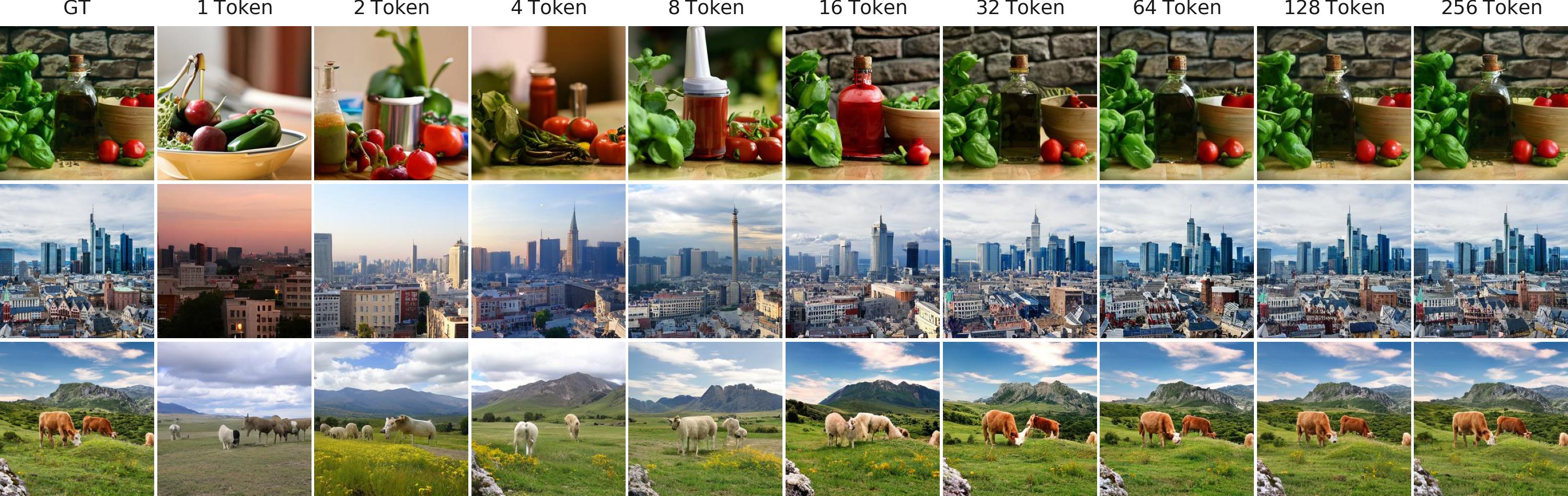}
\caption{ 
Image reconstruction using our structured visual tokenization approach, which uniquely enables decoding at any token count. Each column shows reconstructions resulting from progressively increasing the number of tokens, from a single token to 256 tokens. 
Unlike conventional tokenizers that require a fixed number of tokens for meaningful decoding, our method ensures that each token incrementally refines the image, with earlier tokens capturing the most salient features and later ones adding finer details. 
This demonstrates the flexibility and effectiveness of our approach in producing coherent images even with very few tokens (view more from \cref{fig:teaser_extend} in the Appendix).
}
\label{fig:teaser}
\vspace{2em}
}]

\maketitle

\addtocontents{toc}{\protect\setcounter{tocdepth}{0}}

\begin{abstract}
We introduce a novel visual tokenization framework that embeds a provable PCA-like structure into the latent token space. 
While existing visual tokenizers primarily optimize for reconstruction fidelity, they often neglect the structural properties of the latent space---a critical factor for both interpretability and downstream tasks.
Our method generates a 1D causal token sequence for images, where each successive token contributes non-overlapping information with mathematically guaranteed decreasing explained variance, analogous to principal component analysis. 
This structural constraint ensures the tokenizer extracts the most salient visual features first, with each subsequent token adding diminishing yet complementary information. 
Additionally, we identified and resolved a semantic-spectrum coupling effect that causes the unwanted entanglement of high-level semantic content and low-level spectral details in the tokens by leveraging a diffusion decoder.
Experiments demonstrate that our approach achieves state-of-the-art reconstruction performance and enables better interpretability to align with the human vision system.
Moreover, autoregressive models trained on our token sequences achieve performance comparable to current state-of-the-art methods while requiring fewer tokens for training and inference.
\end{abstract}

\section{Introduction}

\begin{flushright}
\textit{The limits of my language mean the limits of my world.}\\\vspace{.2em}
---Ludwig Wittgenstein\\\vspace{.2em}
\textit{Tractatus Logico-Philosophicus}
\end{flushright}

\begin{figure*}[t]
\centering
\begin{subfigure}[b]{0.77\textwidth}
    \centering
    \includegraphics[width=\linewidth]{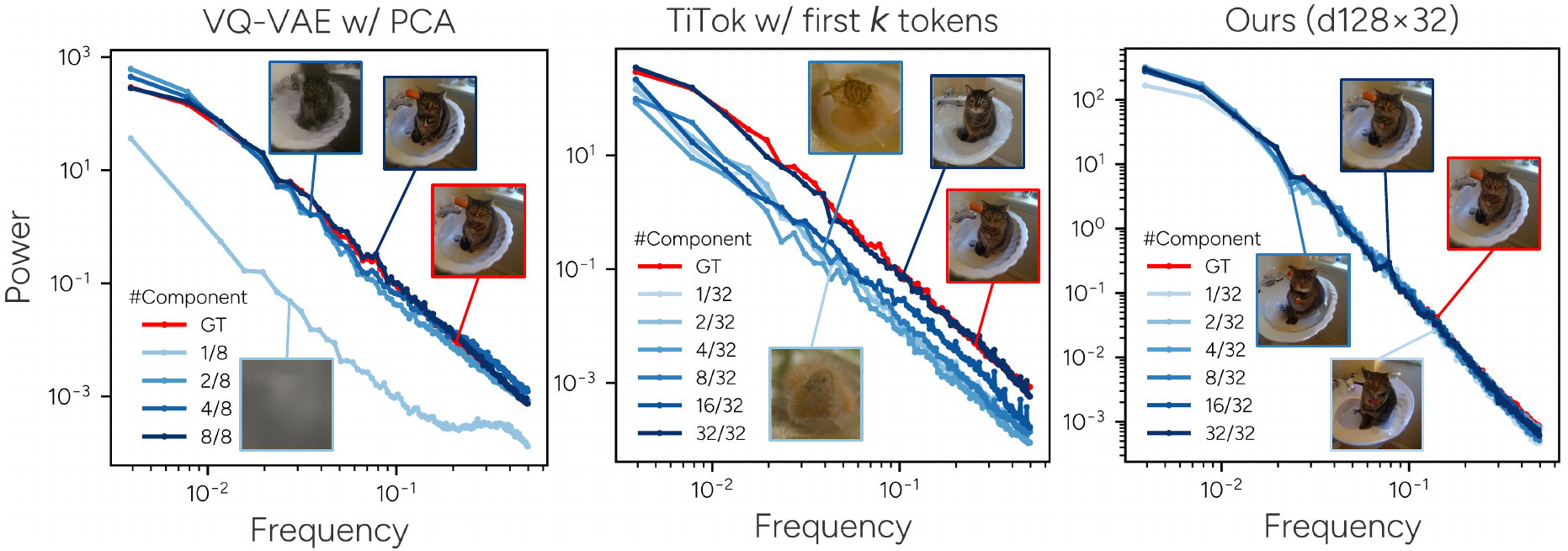}
    \caption{
    \textbf{Semantic-spectrum coupling.} Comparison of the frequency-power spectra for different tokenizers. Here, we decompose the tokens from the tokenizers to demonstrate their contribution to the spectrum of the generated image. 
    The VQ-VAE tokenizer~\cite{llamagen} is decomposed by performing PCA in its latent token space, and the 1D TiTok~\cite{titok} is decomposed by replacing all but the first $k$ tokens with a mean token.
    For \modelname, on the other hand, we can clearly see that with any number of tokens, the spectrum remains closely matched with the original image, demonstrating that \modelname can decouple semantics and spectrum in its tokenization process.
    }
    \label{fig:spectral_analysis}
\end{subfigure}\hfill
\begin{subfigure}[b]{0.215\textwidth}
    \centering
    \includegraphics[width=\linewidth]{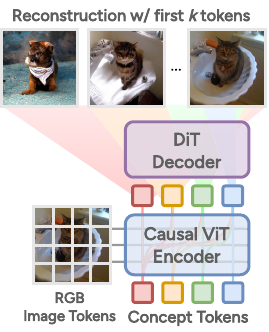}
    \caption{Our tokenizer decomposes the image into visual concepts following a PCA-like coarse-to-fine structure where first few tokens capture most semantic information and the rest refine the details.}
    \label{fig:PCA_structure}
\end{subfigure}
\caption{Spectrum analysis and the PCA-like structure of our tokenizer.}
\end{figure*}

The pursuit of compact visual representations has long been a fundamental goal, driving advancements in visual recognition~\cite{turk1991face,hinton2006reducing} and image generation~\cite{titok,VQVAE}. 
One of the earliest approaches, Principal Component Analysis (PCA)~\cite{shlens2014tutorial}, achieves this by introducing decorrelated, orthogonal components that capture the most significant variations in the data in a progressively diminishing manner (i.e., orderliness), thereby reducing redundancies. 
This enables PCA to effectively reduce dimensionality while preserving essential information, making it a powerful tool for compact representation. 
Building on this foundation, Hinton and Salakhutdinov~\cite{hinton2006reducing} proposed a nonlinear generalization of PCA using autoencoders, which further emphasizes a structured latent space for effective learning and reconstruction error minimization.

While modern approaches such as (vector quantized) variational autoencoders~\cite{VQVAE,vae} share similar goals as earlier methods---compressing images into a compact, low-dimensional space while minimizing reconstruction errors---they have largely abandoned the inherent structural properties, such as orthogonality and orderliness, that were critical to the success of earlier PCA-based techniques.
For instance, mainstream methods employ a 2D latent space~\cite{llamagen,vae,VQVAE}, where image patches are encoded into latent vectors arranged in a 2D grid. While this approach achieves high reconstruction quality, it introduces redundancies that scale poorly as image resolution increases. More recently, 1D tokenizers~\cite{titok,alit} have been proposed to find a more compact set of latent codes for image representation. Although these methods more closely resemble earlier compression approaches~\cite{turk1991face,hinton2006reducing}, they lack structural constraints on latent vectors, making optimization challenging and often resulting in high reconstruction errors. %

We further investigate the latent space of state-of-the-art methods, including VQ-VAE~\cite{llamagen} and TiTok~\cite{titok}, and find that the lack of a structured latent space leads to an inherent tendency for their learned representations to couple significant semantic-level content with less significant low-level spectral information---a phenomenon we refer to as \textit{semantic-spectrum coupling}.
As shown in~\cref{fig:spectral_analysis}, increasing the number of latent codes simultaneously affects both the power spectrum (reflecting low-level intensity information) and the reconstruction of semantic content in the image. Further details on this coupling effect are presented in~\cref{fig:ss_coupling}.

The above motivates us to ask: Can insights from classic PCA techniques be integrated with modern 1D tokenizers to achieve a compact, structured representation of images---one that reduces redundancy while effectively decoupling \textit{semantic} information from less important low-level details?

To this end, we reintroduce a PCA-like structure---incorporating both orthogonality and orderliness---into 1D latent tokens.
Specifically, we propose a dynamic nested classifier-free guidance (CFG) strategy during training to induce an orderliness bias in the tokens, enforcing the emergence of a PCA-like structure where token importance progressively decreases. This is achieved by incrementally replacing later tokens in the 1D sequence with a null condition token at an increasing probability, thereby encouraging earlier tokens to capture the most semantically significant features. This strategy also implicitly promotes orthogonal contributions among tokens (see~\cref{sec:PCA_proof}).
By doing so, our approach ensures a coarse-to-fine token hierarchy with decreasing importance, where each token contributes unique information to reconstruct the image. 
This stands in contrast to previous 1D tokenizers~\cite{titok,alit}, which enforce a 1D structure but lack the orderliness and orthogonality properties that our method introduces (see ~\cref{fig:spectral_analysis}).  
Moreover, the PCA-like structural property enables a flexible encoding of the image by using the most significant tokens.

However, inducing a PCA-like structure alone is insufficient. The visual world exists in a high-dimensional space, and the nested CFG technique might converge to an arbitrary PCA-like structure that does not necessarily disentangle semantically significant content from less important low-level details.
To ensure semantically meaningful features emerge in the latent tokens, we propose leveraging a diffusion-based decoder that follows a spectral autoregressive process~\cite{diffusion_is_spectral_ar,rissanen2022generative}, which progressively reconstructs images from low to high frequencies, conditioned on our 1D latent codes. By doing so, the 1D latent tokens are encouraged to focus on capturing semantically significant information while avoiding entanglement with low-level spectral information.

Finally, the 1D latent codes derived from our tokenizer exhibit a PCA-like hierarchical structure with progressively diminishing significance while also being semantically meaningful.
As shown in \cref{fig:teaser}, all reconstructed images retain the semantic essence of the original, while just 64 tokens are sufficient for high-quality reconstruction. %
Moreover, as illustrated in \cref{fig:spectral_analysis}, the power spectrum profile remains nearly identical to that of the original image, regardless of the number of tokens. This suggests that the latent tokens effectively capture semantically significant information while avoiding entanglement with low-level spectral details.
Notably, the coarse-to-fine structure of latent tokens mirrors the global precedence effect in human vision~\cite{NAVON1977353,fei2007we}, a phenomenon corroborated by our human perceptual evaluations in \cref{sec:human_eval}.

In our experiments, we demonstrate that \modelname can achieve state-of-the-art (SOTA) reconstruction FID scores~\cite{fid} on the ImageNet validation set, surpassing the previous SOTA tokenizer by almost 10\% in FID.
\modelname maintains this SOTA reconstruction performance while maintaining a compact latent suitable for generative modeling.
The autoregressive model trained on the tokens generated by \modelname can achieve comparable performance with the SOTA models while requiring only 32 tokens for training and inference.
Additionally, linear probing in the latent space generated by \modelname also performs up to 63.5\% top-1 classification accuracy, indicating \modelname can capture not only the essence for reconstructing high fidelity images but also the linear separable features.

\section{Related Work}

\noindent \textbf{Image tokenization} aims to transform images into a set of compact latent tokens that reduces the computation complexity for generative models like diffusion models and autoregressive models.
Thus, we view image tokenization as a way of decomposing images into a learnable ``language" for the generative model.
VQ-VAE~\cite{VQVAE} is among the most widely used visual tokenizers. Combining vector quantization into the VAE~\cite{vae} framework, VQ-VAE can generate discrete tokens for the input image.
Improvements over the VQ-VAE have also been proposed, such as VQGAN~\cite{VQGAN}, which introduces adversarial loss to improve the reconstruction quality, and RQ-VAE~\cite{rqvae}, which introduces multiple vector quantization stages.
The insufficient usage of a codebook for vector quantization in VQ-VAE has also raised issues, and MAGVIT-v2~\cite{magvitv2} introduces Look-up Free Quantization (LFQ) to alleviate this issue. 
Semantic information from a pre-trained visual foundation model~\cite{radford2021clip,MAE,dinov2} has also been shown to be beneficial for improving codebook usage~\cite{vqganlc}, improving reconstruction~\cite{lightningdit}, and enabling better generation by diffusion models~\cite{MAETok}.
Maskbit~\cite{maskbit} proposed a modernized VQGAN framework with a novel binary quantized token mechanism that enables state-of-the-art conditional image generation performance.
Most recently, \cite{vitok} demonstrates the scaling laws for ViT-based~\cite{vit} visual tokenizers trained with perceptual and adversarial losses.

Though effective, these models tokenize the image into a 2-dimensional array of tokens where there is no obvious way of performing causal autoregressive modeling.
TiTok~\cite{titok} and SEED~\cite{SEED} are among the first works to introduce tokenizers that generate tokens with a 1-dimensional causal dependency. 
This dependency enables large language models to understand and generate images.
The causal ordering has also been shown to be especially helpful for autoregressive generative models~\cite{CRT}.
VAR~\cite{var} takes another approach by formulating visual autoregressive modeling as a next-scale prediction problem.
A multi-scale residual quantization (MSRQ) tokenizer is proposed in~\cite{var} which tokenizes the image in a low-to-high resolution fashion.
Following the development of the 1D tokenizers, ALIT~\cite{alit} demonstrates that 1D tokens can be made adaptive to the image content, leading to an adaptive length tokenizer that can adjust its token length by considering the image entropy, familiarity, and downstream tasks.
However, due to the semantic-spectrum coupling effect within these 1D causal tokenizers, the structure within the token sequence generated by these tokenizers is still not clear.
In this work, we introduce a novel tokenizer that is able to encode an image to a 1D causal sequence with provable PCA-like structures. 
By decoupling the semantic features and the spectrum information with a diffusion decoder, this tokenizer is not only useful for vision generative modeling but also closely aligns with human perception and enables better interpretability and downstream performance.
Concurrent works~\cite{onedpiece,bachmann2025flextok} also employed similar dropout techniques as ours to form a variable-length 1D tokenizer.
We refer interested readers on this topic to this fascinating blog that introduces many concepts in tokenization~\cite{dieleman2025latents}.

\noindent \textbf{Modern generative vision modeling}
can be roughly divided to two categories, diffusion-based modeling~\cite{dit} and autoregressive-based modeling~\cite{llamagen}. 
Both modeling techniques typically require a visual tokenizer to compress the input visual signal to a compact space for efficient learning.
Diffusion models demonstrate a strong performance since it was introduced~\cite{DDPM}. They typically follow an iterative refinement process, which gradually denoises from a noisy image sampled from a Gaussian distribution to a clean image.
Developments have made efforts toward sharper sample generation~\cite{DDPM,dhariwal2021diffusion}, and faster generation~\cite{ddim}.
The key development in diffusion models is the introduction of latent diffusion models~\cite{LDM}, which allows the diffusion process to be performed in the latent space of a tokenizer~\cite{VQGAN}.
This drastically reduces the computation cost of the diffusion process and enables many more applications~\cite{dit,dalle2}.
Moreover, theoretical understanding of diffusion models has shown that the denoising process can be roughly described as a spectral autoregressive process~\cite{diffusion_is_spectral_ar,rissanen2022generative} where the model uses all previously seen lower frequency information to generate higher frequency information.
Autoregressive models have also been developed for vision modeling, they typically follow a left-to-right generation process where the model predicts the next pixel given the previous pixels~\cite{pixelcnn,pixelrnn}.
Recent works have been developing more advanced autoregressive models leveraging the architecture improvements from NLP and advanced image tokenization~\cite{llamagen}.
Autoregressive models would naturally require an order for which to generate the tokens, some works apply random masking to allow the model to learn random order~\cite{RAR,mar,maskgit}.
VAR~\cite{var} introduces a novel paradigm of next-scale-prediction formulation for visual autoregressive modeling, introducing a natural order---scale---to autoregressive modeling.
In this work, in observation of the semantic spectrum coupling, we leverage diffusion as a decoder for our tokenizer for its spectral auto-regression property to decouple semantic from spectrum information.
Additionally, in our experiments, we demonstrate that we can train autoregressive models on the tokens generated by our tokenizers to achieve a comparable performance with the state-of-the-art models.

\section{Preliminary}
\label{sec:diffusion_prelim}
In this section, we provide a concise summary of the denoising diffusion model~\cite{DDPM} as a preliminary for understanding our \modelname architecture.

\subsection{Denoising Diffusion Models}

A $T$-step Denoising Diffusion Probabilistic Model (DDPM)~\cite{DDPM} consists of two processes: the forward process (also referred to as diffusion process), and the reverse inference process. 
The forward process from data $\bm{x}_0 \sim q_{\text{data}}(\bm{x}_0)$ to the latent variable $\bm{x}_T$ can be formulated as a fixed Markov chain: $q(\bm{x}_1, ..., \bm{x}_T | \bm{x}_0) = \prod_{t=1}^T q(\bm{x}_t | \bm{x}_{t-1})$,
where $q(\bm{x}_t | \bm{x}_{t-1}) = \mathcal{N}(\bm{x}_t; \sqrt{1-\beta_t}\bm{x}_{t-1}, \beta_t \bm{I})$ is a normal distribution, $\beta_t$ is a small positive constant. The forward process gradually perturbs $\bm{x}_0$ to a latent variable with an isotropic Gaussian distribution $p_{\text{latent}}(\bm{x}_T) = \mathcal{N}(\mathbf{0}, \bm{I})$. 

The reverse process strives to predict the original data $\bm{x}_0$ from the latent variable $\bm{x}_T \sim \mathcal{N}(\mathbf{0}, \bm{I})$ through another Markov chain:
$p_{\theta}(\bm{x}_0, ..., \bm{x}_{T-1} | \bm{x}_T) =  \prod_{t=1}^T p_{\theta}(\bm{x}_{t-1} | \bm{x}_t)$.
The training objective of DDPM is to optimize the Evidence Lower Bound~(ELBO):
$\mathcal{L}=\mathbb{E}_{\bm{x}_0, \bm{\epsilon}}||\bm{\epsilon}-\bm{\epsilon}_{\theta}(\bm{x}_t, t)||^2_2,$ 
where $\bm{\epsilon}$ is the Gaussian noise in $\bm{x}_t$ which is equivalent to $\triangledown_{\bm{x}_t}  \ln{q(\bm{x}_t | \bm{x}_0)} $, $\bm{\epsilon}_{\theta}$ is the model trained to estimate $\bm{\epsilon}$.
Conditional diffusion models~\cite{LDM} maintain the forward process and directly inject the condition $\bm{z}$ into the training objective:
\begin{equation}
\mathcal{L}=\mathbb{E}_{\bm{x}_0, \bm{\epsilon}}||\bm{\epsilon}-\bm{\epsilon}_{\theta}(\bm{x}_t, \bm{z}, t)||^2_2 \,, \nonumber
\end{equation}
where $\bm{z}$ is the condition for generating an image with specific semantics.
Except for the conditioning mechanism, the Latent Diffusion Model (LDM)~\cite{LDM} takes the diffusion and inference processes in the latent space of VQGAN~\cite{VQGAN}, which is proven to be more efficient and generalizable than operating on the original image pixels.

\begin{figure}[t]
    \centering
    \includegraphics[width=\linewidth]{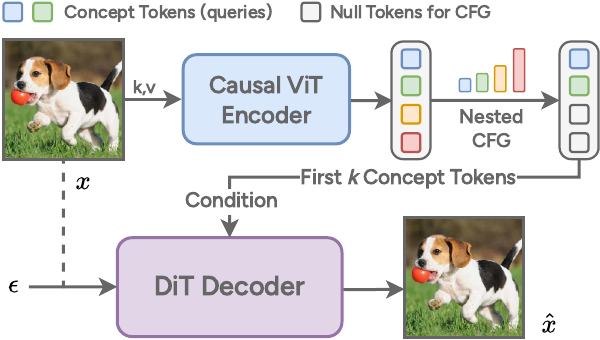}
    \caption{
    \modelname tokenizer architecture. 
    The ViT encoder resamples the 2D image patch tokens into a 1D causal sequence of concept tokens. 
    These concept tokens are then used as conditions to the DiT decoder to reconstruct the original image.
    To induce a PCA-like structure in the concept tokens, we apply nested CFG.}
    \label{fig:arch}
\end{figure}

\section{\modelname Architecture}

\subsection{Overview}
The design of \modelname aims to generate a compact latent code representation of the visual content with a mathematically-guaranteed structure.
As a first step to induce a PCA-like structure, we need to be able to encode the images to a sequence with 1D causal ordering.
Thus we require an encoder $E:\mathbbm{R}^{H\times W\times C}\rightarrow\mathbbm{R}^{K\times D}$ to map the input image of shape $H\times W \times C$ to $K$ causally ordered tokens, each with a dimension of $D$.
We leverage the vision transformer~\cite{vit} (ViT) architecture to implement this encoder for \modelname.
Specifically, we first encode the input image $\bm{x}_0\in\mathbbm{R}^{H\times W\times C}$ to a sequence of image patches $\bm{X}_{\text{patch}}$, we also randomly initialize a set of concept tokens $\bm{X}_{\text{concept}}=\{\bm{z}_1, \bm{z}_2, \dots, \bm{z}_K\}$ to be passed into the transformer model.
Thus, the transformer model within the encoder $E$ takes the below concatenated token sequence for processing:
\begin{equation}
\bm{X}=\lbrack\bm{X}_{\text{cls}};\bm{X}_{\text{patch}};\bm{X}_{\text{concept}}\rbrack \,, \nonumber
\end{equation}
where $\bm{X}_{\text{cls}}$ is the \texttt{[CLS]} token. For the patch tokens $\bm{X}_{\text{patch}}$ and the \texttt{[CLS]} we do not perform any masking to mimic the standard ViT behavior and for the concept tokens $\bm{X}_{\text{concept}}$, we apply a causal attention masking such that only preceding tokens are visible (as illustrated in \cref{fig:PCA_structure}) to enforce them learn a causally ordered tokenization.
After the ViT has processed the tokens, the output concept tokens $\bm{X}_{\text{concept}}=\{\bm{z}_1, \bm{z}_2, \dots, \bm{z}_K\}$ are used as the condition input to the diffusion-based decoder $D:\mathbbm{R}^{K\times D}\times\mathbbm{R}^{H\times W\times C}\rightarrow\mathbbm{R}^{H\times W\times C}$ to learn to denoise a noisy version $\bm{x}_t$ of the input image $\bm{x}_0$.
Doing this alone would allow the encoder $E$ to encode the information about the image into the concept tokens $\bm{X}_{\text{concept}}$.
However, this information is not structured.
To induce a PCA-like structure with the concept tokens, we apply the nested CFG technique $N:\mathbbm{R}^{K\times D}\rightarrow\mathbbm{R}^{K\times D}$ that will be introduced in later sections.
To summarize, the training process of \modelname is to minimize this training loss similar to the training of a conditional diffusion model:
\begin{equation}
\mathcal{L}=\mathbb{E}_{\bm{x}_0, \bm{\epsilon}}||\bm{\epsilon}-D(\bm{x}_t, N(E(\bm{X})), t)||^2_2 \,, \nonumber
\end{equation}
where $t$ is the condition for the forward process timestep, and $\bm{\epsilon}$ is the noise at timestep $t$.
Note that \modelname generates continuous tokens instead of quantized tokens like previous works~\cite{llamagen,titok}.
The reason is that we hope \modelname to capture the PCA-like variance decay structure, which is hard to capture when using quantized tokens. 
With the usage of Diffusion Loss~\cite{mar}, our experiments on autoregressive modeling with continuous tokens have shown that this design does not affect generative modeling performance.

\subsection{Diffusion-based Decoder}
The decoder for \modelname is based on the conditional denoising diffusion model.
The decoder is implemented using the Diffusion-Transformer (DiT) architecture~\cite{dit}, with the condition $\bm{X}_{\text{concept}}$ injected by cross-attention.
For efficient training, we adopt the LDM technique by training the decoder on the latent space of a pretrained VAE model~\cite{LDM}.
This design choice stems from the observation of the semantic-spectrum coupling effect. %
Whereas, if a deterministic decoder is used to directly regress the pixel values like previous state-of-the-art visual tokenizers, the token space learned by the encoder $E$ will entangle the semantic content and the spectral information of the image.
This will prevent the encoder $E$ from learning a semantic meaningful PCA-like structure where the first tokens capture the most important semantic contents.
In~\cref{sec:diffusion_prelim}, we describe the diffusion forward process as gradually corrupting the image with Gaussian noise, which is filtering out more and more high-frequency information.
Since the training objective for the diffusion model is to reverse this forward process, the model naturally learns to generate low-frequency information first and then high-frequency details.
This is described in~\cite{diffae,rissanen2022generative} as a spectral autoregressive process where the diffusion process itself can already generate the spectral information, leaving the conditions $\bm{X}_{\text{concept}}$ to be able to encode most semantic information rather than spectral information.

\subsection{Inducing the PCA-like Structure}
\label{sec:prove_pca}

Although the encoder \(E\) and diffusion-based decoder \(D\) can produce high-quality reconstructions, their concept tokens \(\bm{X}_{\text{concept}}\) lack any explicit structural regularization beyond causal ordering. 
To impose a hierarchical variance-decaying property similar to PCA~\cite{shlens2014tutorial}, we introduce a \textbf{nested classifier-free guidance (CFG)} function  
\begin{equation}
N:\mathbb{R}^{K \times D} \times \{1, \dots, K\} \to \mathbb{R}^{K \times D} \,, \nonumber
\end{equation}
inspired by nested dropout~\cite{nested_dropout,iob} and Matryoshka representations~\cite{kusupati2022matryoshka}.
Specifically, we sample an integer \(k' \sim \mathcal{U}\{1, \dots, K\}\) and apply \(N\) to the concept tokens \(\bm{X}_{\text{concept}} = (\bm{z}_1,\dots,\bm{z}_K)\) as follows:  
\[
N(\bm{X}_{\text{concept}}, k') \;=\; (\bm{z}_1,\dots,\bm{z}_{k'-1}, \bm{z}_{\emptyset}, \dots, \bm{z}_{\emptyset}) \,,
\]  
where \(\bm{z}_{\emptyset}\) are learnable \textit{null-condition} tokens for masked positions. 
Intuitively, forcing positions \(k', \dots, K\) to become null tokens compels the \textbf{earlier tokens} \(\bm{z}_1,\dots,\bm{z}_{k'-1}\) to encode the most salient semantic content. 
Over the course of training, the uniform sampling of \(k'\) induces a coarse-to-fine hierarchy in the concept tokens, mirroring the variance-decaying property of PCA~\cite{shlens2014tutorial}.  

From a classifier-free guidance~\cite{cfg} perspective, each token \(\bm{z}_k\) can be viewed as a conditional signal, and applying \(N\) effectively provides an “unconditional” pathway for later tokens. 
In \cref{sec:PCA_proof}, we formally show that this procedure yields a PCA-like structure in which the earliest tokens capture the largest share of variance. A high-level illustration of our overall architecture is provided in \cref{fig:arch}.

\begin{table*}[t]
\centering 
\tablestyle{5.5pt}{1.05}
\begin{tabular}{l ccc |ccc | cccccc}
    Method & \#Token & Dim. & VQ & rFID$\downarrow$ & PSNR$\uparrow$ & SSIM$\uparrow$ & Gen. Model & Type & \#Token & \#Step & gFID$\downarrow$ & IS$\uparrow$ \\
    \shline
    MaskBit~\cite{maskbit} & 256 & 12 & \cmark & 1.61 & -- & -- & MaskBit & Mask. & 256 & 256 & 1.52 & 328.6 \\
    RCG (cond.)~\cite{RCG} & 1 & 256 & \xmark & -- & -- & -- & MAGE-L & Mask. & 1 & 20 & 3.49 & 215.5 \\
    MAR~\cite{mar} & 256 & 16 & \xmark & 1.22 & -- & -- & MAR-L & Mask. & 256 & 64 & 1.78 & 296.0 \\
    TiTok-S-128~\cite{titok} & 128 & 16 & \cmark & 1.71 & -- & -- & MaskGIT-L & Mask. & 128 & 64 & 1.97 & 281.8 \\
    TiTok-L-32~\cite{titok} & 32 & 8 & \cmark & 2.21 & -- & -- & MaskGIT-L & Mask. & 32 & 8 & 2.77 & 194.0 \\
    \hline
    VQGAN~\cite{VQGAN} & 256 & 16 & \cmark & 7.94 & -- & -- & Tam. Trans. & AR & 256 & 256 & 5.20 & 280.3 \\
    ViT-VQGAN~\cite{vitvqgan} & 1024 & 32 & \cmark & 1.28 & -- & -- & VIM-L & AR & 1024 & 1024 & 4.17 & 175.1  \\
    RQ-VAE~\cite{rqvae} & 256 & 256 & \cmark & 3.20 & -- & -- & RQ-Trans. & AR & 256 & 64 & 3.80 & 323.7 \\
    VAR~\cite{var}  & 680 & 32 & \cmark & 0.90 & -- & -- & VAR-$d$16 & VAR & 680 & 10 & 3.30 & 274.4 \\
    ImageFolder~\cite{imagefolder} & 286 & 32 & \cmark & 0.80 & -- & -- & VAR-$d$16 & VAR & 286 & 10 & 2.60 & 295.0\\
    LlamaGen~\cite{llamagen} & 256 & 8 & \cmark & 2.19 & 20.79 & 0.675 & LlamaGen-L & AR & 256 & 256 & 3.80 & 248.3 \\
    CRT~\cite{CRT} & 256 & 8 & \cmark & 2.36 & -- & -- & LlamaGen-L & AR & 256 & 256 & 2.75 & 265.2 \\
    Causal MAR~\cite{mar} & 256 & 16 & \xmark & 1.22 & -- & -- & MAR-L & AR & 256 & 256 & 4.07 & 232.4 \\
    \hline
    \modelname w/ DiT-L & 256 & 16 & \xmark &  0.78  & 21.61 & 0.626 & \armodelname-L & AR & 32 & 32 & 2.57 & 260.9  \\
    \modelname w/ DiT-XL & 256 & 16 & \xmark &  0.72 & 21.43 & 0.613 & \armodelname-L & AR & 32 & 32 & 2.57 & 254.0 \\
\end{tabular}
\caption{Reconstruction and generation performance on ImageNet. ``Dim." denotes the dimension of the tokens, and ``\#Step" denotes the number of steps needed for generating the complete image. ``\#Token'' stands for the number of tokens used for image reconstruction (left) and generation (right), respectively.}%
\label{tab:tok_comp}
\end{table*}

\subsection{Autoregressive Modeling with \bf\modelname}
With the learned latent token sequences $\bm{X}_{\text{concept}}=\{\bm{z}_1, \bm{z}_2, \dots, \bm{z}_K\}$ obtained from a well-trained encoder, we can train an autoregressive model for image generation.
Specifically, we leverage the architecture of LlamaGen~\cite{llamagen} for autoregressive modeling, which is a modern variant of GPT~\cite{radford2018improving} where pre-norm~\cite{prenorm} is applied with RMSNorm~\cite{rmsnorm}, and SwiGLU~\cite{swiglu} activation function is used.
As \modelname adopts continuous tokens, the prediction head of our autoregressive model is a denoising MLP following MAR~\cite{mar} that is supervised by the diffusion process~\cite{DDPM}.
Specifically, the autoregressive model is modeling a next token prediction problem of $p(\bm{z}_k|\bm{z}_{<k}, c)=G_{\text{discrete}}(\bm{z}_{<k}, c),$ where $c$ is the class label embedding and $G_{\text{discrete}}$ being the causal transformer to predict the next token with all previous tokens.
If the latent tokens generated by $E$ were quantized, we can directly leverage the softmax prediction head to obtain the next token prediction~\cite{llamagen}.
However, $E$ generates continuous tokens. Thus we leverage the design from~\cite{mar} to instead predict a condition $\bm{m}_k$ from all previous tokens $\bm{z}_{<k}$ and $c$, $\bm{m}_k = G(\bm{z}_{<k}, c)$.
This condition $\bm{m}_k$ is used to condition a small diffusion MLP model to generate the $k$-th token $\bm{z}_k$ from noise $\bm{z}_k^T$.
Specifically, the autoregressive model $G$ and the diffusion MLP model $M$ is trained with a similar diffusion loss as defined in~\cref{sec:diffusion_prelim}:
\begin{equation}
    \mathcal{L}_{\text{G}} = \mathbb{E}_{\bm{z}_k^0, \bm{\epsilon}}||\bm{\epsilon}-M(\bm{z}_k^t, G(\bm{z}_{<k}, c), t)||^2_2 \,, \nonumber
\end{equation}
where $\bm{z}_k^0$ is the ground truth next token $\bm{z}_k$.
This continuous autoregressive modeling design enables the usage of the continuous tokens from \modelname to perform generative autoregressive modeling with comparable performance to the current state-of-the-art generative models.
As this model uses the noise $\bm{\epsilon}$ as a learning objective, we term this generative model as \armodelname.

\section{Experiments}

Following the common practice~\cite{mar}, we experiment on ImageNet~\cite{imagenet} at a resolution of 256$\times$256, and report FID~\cite{fid} and IS~\cite{is} tested with the evaluation suite provided by~\cite{dhariwal2021diffusion}.

\subsection{Implementation Details}

\paragraph{\modelname autoencoder.}
The encoder of \modelname is a standard ViT-B/16~\cite{vit}, except for additional concept tokens and causal attention masks applied to them.
We fix the size (token count$\times$ dimension) of concept tokens representing an image to 4096 and consider four variants with token dimensions ranging from 16 to 256. 
Unless otherwise specified, we use 16-dimensional concept tokens (denoted as d16$\times$256) by default, which are more friendly for reconstruction as shown in~\cref{fig:recon_16dim_scale} in the Appendix.
Before being fed to the decoder, the concept tokens are normalized by their own mean and variance following~\cite{RCG}.
The decoder is a DiT~\cite{dit} with a patch size of 2. 
We experiment across different scales (B, L, and XL) and take DiT-L as default. 
The decoder operates on the latent space of a publicly available KL-16 VAE provided by~\cite{mar} to reduce computation cost. 
The VAE is frozen during training, and both the encoder and decoder are trained from scratch.
To enforce the quality of the learned concept tokens and stabilize training, we apply REPA~\cite{repa} with DINOv2-B~\cite{dinov2} as a regularizer to the 8th layer of the DiT decoder.

\paragraph{Autoregressive image modeling.}
We validate the effectiveness of \modelname autoencoder by training autoregressive image generation models using LlamaGen~\cite{llamagen} combined with diffusion loss~\cite{mar}. 
The input sequence is pre-pended with a \texttt{[CLS]} token for class conditioning, which is randomly dropped out with probability 0.1 during training for classifier-free guidance. 
At inference time, we use a CFG schedule following~\cite{mar,muse}, and do not apply temperature sampling. 
Note that the implementation can be general. 
While our preliminary validation demonstrates promising results, we anticipate better configurations in future work.

\subsection{Reconstruction and Generation Quality}

\cref{tab:tok_comp} presents the comparison of \modelname to state-of-the-art image tokenizers and accompanying generative models.
The comparisons on the image reconstructions are made with the variant of the state-of-the-art models with similar-sized latent space (i.e., token count $\times$ token dimension---the number of floating point numbers used).
\modelname demonstrates a superior reconstruction performance in terms of the rFID score compared to all previous works.
Advancing the state-of-the-art performance in image reconstruction by 10\% in rFID score compared to the next best model of ImageFolder~\cite{imagefolder} with a more compact latent space (286$\times$32 for \cite{imagefolder} and 256$\times$16 for ours).

In terms of generative modeling, the results in~\cref{tab:tok_comp} demonstrate that \modelname can obtain a gFID score comparable to the state-of-the-art tokenizers for standard autoregressive (AR) modeling.
Remarkably, the unique PCA-like structure within the latent space of \modelname enables efficient generative modeling with fewer tokens.
Specifically, \armodelname only requires being trained and evaluated on the first 32 tokens from \modelname to achieve a gFID score comparable with the state-of-the-art generative models on ImageNet.
This efficiency in the number of tokens used for generative modeling would allow \armodelname to use a much smaller number of inference steps to achieve better results.

\begin{figure}
    \centering
    \includegraphics[width=.9\linewidth]{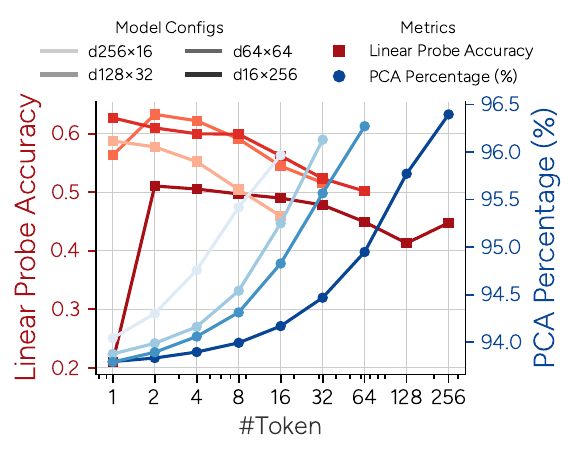}
    \caption{The explained variance ratio from \modelname's PCA-like structure and the linear probing accuracy on the tokens.}
    \label{fig:pca_acc}
\end{figure}

\subsection{Representation and Structure}

We study the property of the structured latent space of \modelname.
In~\cref{fig:pca_acc}, we first generate a PCA-like variance explained plot by varying the number of tokens used to compute the averaged diffusion loss over the validation set and all diffusion timesteps.
After obtaining the averaged loss, the loss value of using all null conditions is treated as the upper bound, and the loss value of using all conditions is treated as the lower bound. We plot the reduction percentage for the contribution of each token to the diffusion loss.
It is clear that \modelname forms a PCA-like structure among the contributions of its tokens.
Furthermore, we perform linear probing on the extracted tokens; the plots are also available in~\cref{fig:pca_acc}.
Comparing this against the variance-explained curve, we can see the linear probing accuracy can reach the highest performance when using a low number of tokens and then gradually decrease as more tokens are used.
This reveals that \modelname tends to store the most salient features (i.e., the object category) in the first few tokens, and thus, they benefit from linear probing. 
And yet, when adding more tokens, details of the scene are added, which causes the linear probing readout to lose category information.

\begin{figure}
    \centering
    \includegraphics[width=\linewidth]{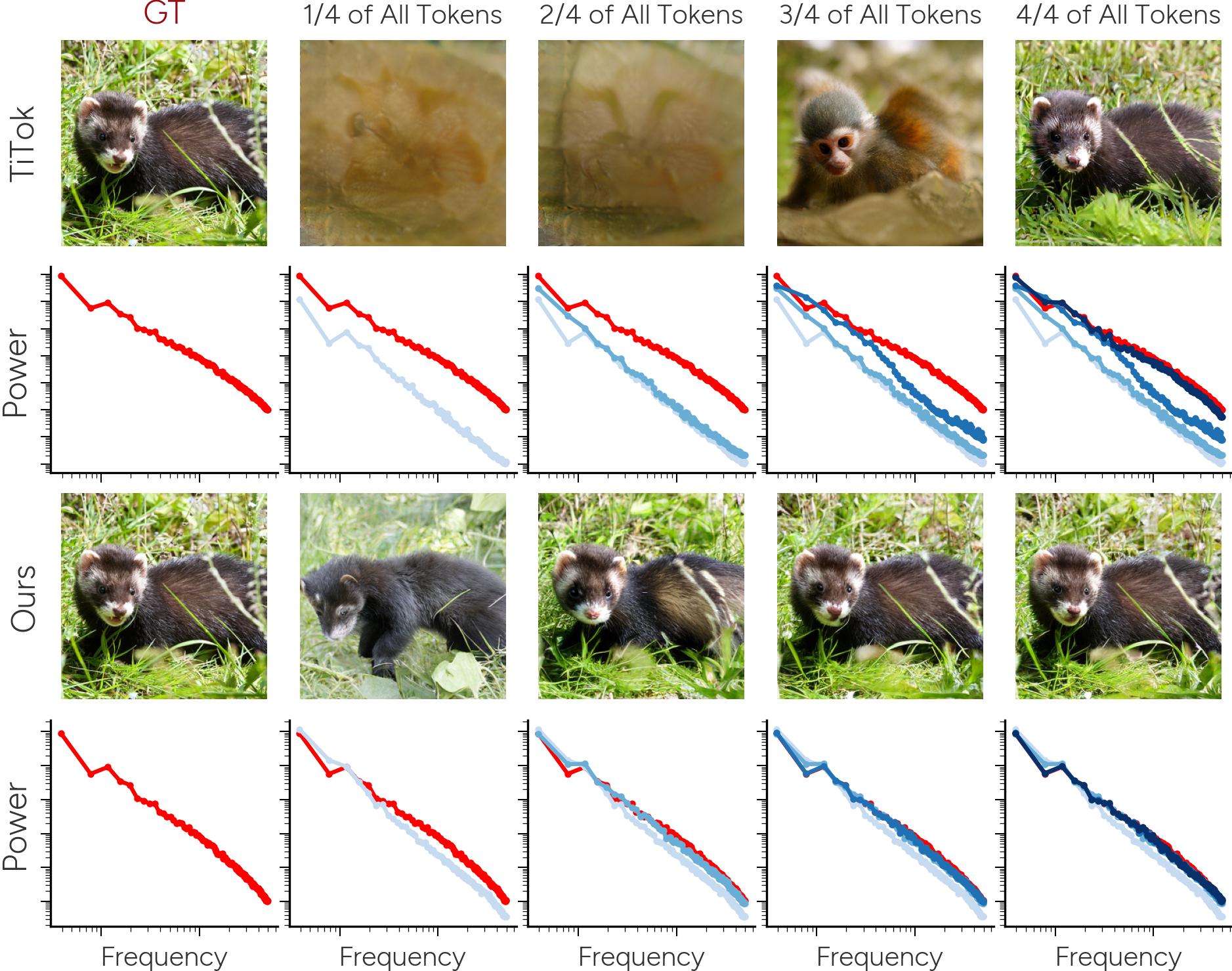}
    \vspace{-.8em}
    \caption{Reconstructed images and their corresponding power-frequency plots, illustrating \textbf{semantic-spectrum coupling}. Each column shows reconstructions using only the first $k$ tokens, increasing from left to right, alongside a plot of the reconstructed image’s frequency power (blue) overlaid on the ground-truth (red) image.}
    \label{fig:ss_coupling}
\end{figure}

\begin{figure}
    \centering
    \includegraphics[width=.8\linewidth]{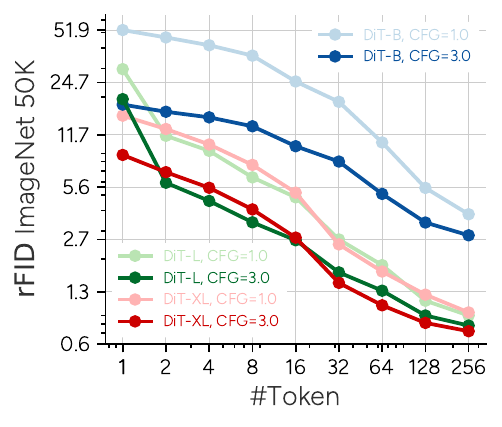}
    \caption{Scaling behavior of different-sized DiT decoder (qualitative results can be found in \cref{fig:recon_16dim_scale} in the Appendix).}
    \label{fig:dit_scale}
\end{figure}

\begin{figure*}[t]
    \centering
    \includegraphics[width=0.85\linewidth]{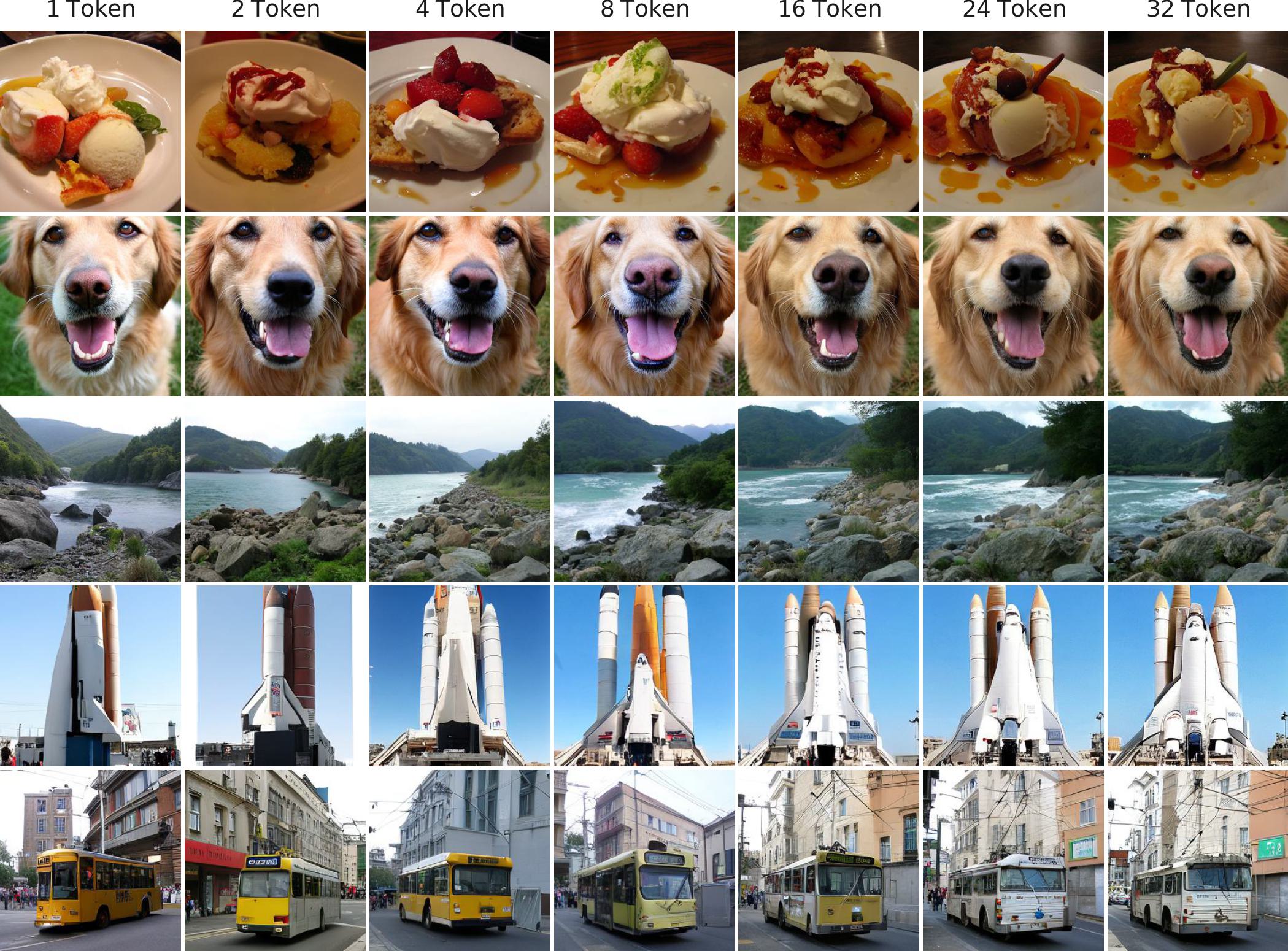}
    \vspace{1em}
    \caption{Examples of the intermediate generation results of \armodelname-L trained on \modelname tokens (see more from \cref{fig:more_example_ar}).}
    \label{fig:example_ar_results}
\end{figure*}

\subsection{Semantic-spectrum Coupling}
We further demonstrate the semantic-spectrum coupling effect in~\cref{fig:ss_coupling}.
The power-frequency plots are obtained by taking the 2D Fourier transform of the image, then averaging the magnitude-squared values over concentric circles in the frequency domain to get a 1D power distribution.
The horizontal axis represents spatial frequency (higher values correspond to finer details), and the vertical axis shows the power at each frequency.
As we can see from~\cref{fig:ss_coupling}, for the TiTok~\cite{titok} model, the semantic content only emerges from using 2/4 to 3/4 of all tokens.
From the power-frequency plot, it is clear that the model can not match the correct power distribution with fewer tokens and that when adding more tokens, not only is the power distribution shifting toward the ground truth distribution, but also, the semantic content emerges.
This is what we term the ``semantic-spectrum coupling" effect, where, when adding more tokens, both semantic content and spectral information are encoded.
On the other hand, it is very clear that \modelname can match the power distribution of the ground truth with only 1/4 of the tokens, and later tokens contribute more to the semantic content of the image, successfully disentangling semantic content from the spectral information.

\subsection{Ablation Study}

We ablate the effect of the scale of the diffusion decoder in \modelname, the number of tokens used for reconstruction, and the strength of classifier-free-guidance in~\cref{fig:dit_scale}.
\modelname follows a very clear scaling behavior in terms of the number of tokens and the model size.
Notably, strengthening the classifier-free-guidance scale can greatly help the reconstruction performance with a smaller model and fewer tokens.
More ablation studies on \armodelname and \modelname are available in~\cref{sec:ablation_ext}.

\subsection{Discussion on Qualitative Results}

In \cref{fig:teaser}, we have shown that \modelname can consistently produce semantic-meaningful high-quality images with any number of tokens that progressively refine towards the reconstruction target. 
Notably, the original image can be represented with as few as 32 tokens. This is especially helpful when we want to re-purpose the tokens for class-conditional image generation, in which the task is to produce semantic-consistent images instead of exact reconstruction. 
Thus, we can train the \armodelname autoregressive model efficiently with only the first few concept tokens (32 in our case).
In \cref{fig:example_ar_results,fig:more_example_ar}, we demonstrate the effectiveness of autoregressive modeling following this strategy. 
The model is a \armodelname-L trained on the first 32 concept tokens of a \modelname (w/ DiT-XL) tokenizer.
It is encouraging to see that the first token the model generates already sketches the majority of the scene well and even generates highly faithful images in easier cases like animals. 
When more tokens are generated, the image is gradually refined and converges to a highly detailed and visually coherent generation.

\section{Conclusion}

We introduce \modelname, a PCA-like structured 1D tokenizer that addresses semantic-spectrum coupling through a dynamic nested classifier-free guidance strategy and a diffusion-based decoder. 
Our method enforces a coarse-to-fine token hierarchy that captures essential semantic features while maintaining a compact latent representation.
Our experiments demonstrate that \modelname achieves state-of-the-art reconstruction FID scores on the ImageNet validation set, surpassing the previous SOTA tokenizer by almost 10\% in FID. Moreover, the autoregressive model \armodelname trained on \modelname's tokens attains comparable performance to current SOTA methods while requiring only 32 tokens for training and inference. Additionally, linear probing in the latent space yields up to 63.5\% top-1 classification accuracy, confirming the effectiveness of our approach in capturing semantic information.
These results highlight the promise of \modelname for high-fidelity image reconstruction and generative modeling, paving the way for more efficient and compact visual representations.

\section*{Acknowledgment}
This work has been supported in part by Hong Kong Research Grant Council---Early Career Scheme (Grant No. 27209621), General Research Fund Scheme (Grant No. 17202422, 17212923), Theme-based Research (Grant No. T45-701/22-R), and Shenzhen Science and Technology Innovation Commission (SGDX20220530111405040). Part of the described research work is conducted in the JC STEM Lab of Robotics for Soft Materials funded by The Hong Kong Jockey Club Charities Trust. 
We sincerely appreciate the dedicated support we received from the participants of the human study. We are also grateful to Anlin Zheng and Haochen Wang for helpful suggestions on the design of technical details.

{
\small
\bibliographystyle{ieeenat_fullname}
\bibliography{main}
}

\input{X_suppl}

\end{document}

%% file: preamble.tex
\usepackage[dvipsnames]{xcolor}

\usepackage{comment}
\usepackage{lipsum}
\usepackage{bm}
\usepackage{microtype}
\usepackage{tabularx,booktabs}

\newcommand{\authorskip}{\hspace{2.5mm}}
\usepackage{pifont}
\newcommand{\cmark}{\ding{51}}%
\newcommand{\xmark}{\ding{55}}%

\newlength\savewidth\newcommand\shline{\noalign{\global\savewidth\arrayrulewidth
  \global\arrayrulewidth 1pt}\hline\noalign{\global\arrayrulewidth\savewidth}}

\newcommand{\tablestyle}[2]{\setlength{\tabcolsep}{#1}\renewcommand{\arraystretch}{#2}\centering\footnotesize}
\definecolor{baselinecolor}{gray}{.9}

\newcolumntype{x}[1]{>{\centering\arraybackslash}p{#1pt}}
\newcolumntype{y}[1]{>{\raggedright\arraybackslash}p{#1pt}}
\newcolumntype{z}[1]{>{\raggedleft\arraybackslash}p{#1pt}}

\usepackage{tocloft}

\makeatletter
\renewcommand{\numberline}[1]{%
  \@cftbsnum #1\@cftasnum\hspace*{1em}\@cftasnumb%
}
\makeatother

%% file: X_suppl.tex
\clearpage
\setcounter{page}{1}
\maketitlesupplementary
\appendix

\addtocontents{toc}{\protect\setcounter{tocdepth}{2}}
{
  \hypersetup{linkcolor=black}
  \tableofcontents
}

\section*{Author Contribution Statement}
X.W. and B.Z. conceived the study and guided its overall direction and planning.
X.W. proposed the original idea of semantically meaningful decomposition for image tokenization.
B.Z. developed the theoretical framework for nested CFG and the semantic spectrum coupling effect and conducted the initial feasibility experiments.
X.W. further refined the model architecture and scaled the study to ImageNet.
B.Z. led the initial draft writing, while X.W. designed the figures and plots.
I.E., J.D., and X.Q. provided valuable feedback on the manuscript.
All authors contributed critical feedback, shaping the research, analysis, and final manuscript.

\section*{Limitations and Broader Impacts}
Our tokenizer contributes to structured visual representation learning, which may benefit image compression, retrieval, and generation. 
However, like other generative models, it could also be misused for deepfake creation, misinformation, or automated content manipulation. Ensuring responsible use and implementing safeguards remains an important consideration for future research.
\modelname also presents several limitations, for example, we employ a diffusion-based decoder, but alternative generative models like flow matching or consistency models could potentially improve efficiency.
Additionally, our framework enforces a PCA-like structure, further refinements, such as adaptive tokenization or hierarchical models, could enhance flexibility.

\section{Proof for PCA-like structure}
\label{sec:PCA_proof}

The conditional denoising diffusion model is using a neural network $\bm{\epsilon}_{\theta}(\bm{x}_t, \bm{z}, t)$ to approximated the score function  $\triangledown_{\bm{x}_t}  \ln{q(\bm{x}_t | \bm{x}_0)} $ which guides the transition from a noised image $\bm{x}_t$ to the clean image $\bm{x}_0$.
For the conditional diffusion decoder in \modelname, the score function can be decomposed as:
\begin{equation}
\bm{\epsilon}_{\theta}(\bm{x}_t, \bm{z}_1, \dots, \bm{z}_k) = \bm{\epsilon}_{\theta}(\bm{x}_t, \emptyset) + \sum_{i=1}^k \gamma_i \Delta \bm{\epsilon}_{\theta}(\bm{x}_t, \bm{z}_i) \,, \nonumber
\end{equation}
where $\emptyset$ is the null condition, $\gamma_i$ is the guidance scale, and $\Delta \bm{\epsilon}_{\theta}(\bm{x}_t, \bm{z}_i)=\bm{\epsilon}_{\theta}(\bm{x}_t, \bm{z}_1, \dots, \bm{z}_i) - \bm{\epsilon}_{\theta}(\bm{x}_t, \bm{z}_1, \dots, \bm{z}_{i-1})$ represents the increment contribution of the concept token condition $\bm{z}_i$ to the score function.
Thus, we can rewrite the diffusion training objective with $k$ conditions with the following:
\begin{equation}
\mathcal{L}_k = \mathbb{E} \biggl[ \Bigl\| \epsilon - \bigl( \epsilon_\theta(\bm{x}_t, \emptyset) + \sum_{i=1}^k \gamma_i \Delta \epsilon_\theta(\bm{x}_t, \bm{z}_i) \bigr) \Bigr\|^2 \biggr] \,. \nonumber
\end{equation}
\paragraph{Orthogonality between contribution of concept tokens.}
At the optimal convergence, the gradient of $\mathcal{L}_k$ w.r.t $\Delta \bm{\epsilon}_{\theta}(\bm{x}_t, \bm{z}_i)$ is zero, thus give us:
\begin{equation}
\begin{split}
\frac{\partial \mathcal{L}_k}{\partial \Delta \epsilon_\theta(\bm{x}_t, \bm{z}_i)} &= \mathbb{E} \Bigl[ \bigl( \epsilon - \bm{\epsilon}_\theta(\bm{x}_t, \emptyset) - \sum_{j=1}^k \gamma_j \Delta\bm{\epsilon}_{\theta}(\bm{x}_t, \bm{z}_j) \bigr) \gamma_i\Bigr]\\&=0 \,. \nonumber
\end{split}
\end{equation}
Since model is at convergence, the residual term $\epsilon - \bm{\epsilon}_\theta(\bm{x}_t, \emptyset) - \sum_{j=1}^k \gamma_j \Delta\bm{\epsilon}_{\theta}(\bm{x}_t, \bm{z}_j)$ can not be further reduced by making further changes to the adjustment from the $i$-th concept token $\Delta\bm{\epsilon}_\theta(\bm{x}_t, \bm{z}_j)$. 
In other words, the residual term and all active conditions $\Delta\bm{\epsilon}_\theta(\bm{x}_t, \bm{z}_j)$ are orthogonal to each other.
Next, we can use induction to prove that at convergence, all $\Delta\bm{\epsilon}_\theta(\bm{x}_t, \bm{z}_j)$ terms are orthogonal to each other similar to PCA.
For the case of $k=1$, we only use one concept token to condition the model, thus we can have:
\begin{equation}
\mathbb{E}\left[ \left( \epsilon -  \bm{\epsilon}_\theta(\bm{x}_t, \emptyset) - \gamma_1 \Delta\bm{\epsilon}_\theta(\bm{x}_t, \bm{z}_1) \right) \Delta\bm{\epsilon}_\theta(\bm{x}_t, \bm{z}_1) \right] = 0 \,. \nonumber
\end{equation}
For the case of $k=2$, for $(i=1,2)$, we have:
\begin{equation}
\begin{split}
\mathbb{E}\biggl[ \Bigl(\epsilon - \bm{\epsilon}_\theta(\bm{x}_t, \emptyset)
 - \sum_{j=1}^2\gamma_j \Delta\bm{\epsilon}_\theta(\bm{x}_t, \bm{z}_j) \Bigr) \Delta\bm{\epsilon}_\theta(\bm{x}_t, \bm{z}_i) \biggr] = 0 \,.  \nonumber
\end{split}
\end{equation}
By substituting the $k=1$ case into this, it can be seen that $\mathbb{E}\left[ \Delta\bm{\epsilon}_\theta(\bm{x}_t, \bm{z}_1)^\top  \Delta\bm{\epsilon}_\theta(\bm{x}_t, \bm{z}_2) \right]=0$.
Assuming this orthogonality holds for the first $k-1$ concept tokens:
$\mathbb{E}\left[ \Delta\bm{\epsilon}_\theta(\bm{x}_t, \bm{z}_i)^\top  \Delta\bm{\epsilon}_\theta(\bm{x}_t, \bm{z}_j) \right]=0\quad \forall i, j < k, i \neq j$.
Then for $i<k$, by substituting
\begin{equation}
\epsilon - \bm{\epsilon}_\theta(\bm{x}_t, \emptyset) = \sum_{j=1}^{k-1}\gamma_j \Delta\bm{\epsilon}_\theta(\bm{x}_t, \bm{z}_j) + \gamma_k \Delta\bm{\epsilon}_\theta(\bm{x}_t, \bm{z}_k) \,, \nonumber
\end{equation}
we can have:
\begin{equation}
\mathbb{E}\left[\Delta\bm{\epsilon}_\theta(\bm{x}_t, \bm{z}_i)^\top \Delta\bm{\epsilon}_\theta(\bm{x}_t, \bm{z}_k)\right] = 0 \,. \nonumber
\end{equation}
Thus, the orthogonality propagates to all pairs $(\bm{z}_i,\bm{z}_k)$ for $i<k$.
By induction, we have orthogonality between all pairs of concept tokens.

\paragraph{Variance Explained Hierarchy.}
Assuming the true noise $\epsilon$ can be reconstructed using the conditional model, we have:
\begin{equation}
\epsilon \approx \bm{\epsilon}_{\theta}(\bm{x}_t, \emptyset) + \sum_{i=1}^k \gamma_i \Delta\bm{\epsilon}_\theta(\bm{x}_t, \bm{z}_i) + \text{residual} \,. \nonumber
\end{equation}
Given the orthogonality of $\Delta\bm{\epsilon}_{\theta}(\bm{x}_t, \bm{z}_i)$ we have proven earlier, the total variance can be decomposed as:
\begin{equation}
\text{Var}(\epsilon)=\sum_{i=1}^k \text{Var}(\gamma_i \Delta\bm{\epsilon}(\bm{x}_t, \bm{z_i})) + \text{Var}(\text{residual}) \,. \nonumber
\end{equation}
Let $\lambda_i = \text{Var}(\gamma_i \Delta\bm{\epsilon}_{\theta}(\bm{x_t}, \bm{z}_i))$, representing the variance explained by concept token condition $\bm{z}_i$.
Our dropout design would have the training objective forces:
\begin{equation}
\lambda_1\geq\lambda_2\geq\dots\geq\lambda_k \,, \nonumber
\end{equation}
as each concept token $\bm{z}_i$ is trained to explain the maximal residual variance after accounting for concept tokens $\bm{z}_1, \dots,\bm{z}_{i-1}$.

Thus, combining the orthogonality and the variance decay, \modelname provably grounds the emergence of a \textbf{PCA-like hierarchical structure} in the learned concept tokens. Providing a simple, effective, and explainable architecture for visual tokenization.

\section{Additional Related Work}

\subsection{Concurrent Related Work}

Concurrent work~\cite{bachmann2025flextok} introduces a 1D tokenizer that focuses on adaptive-length tokenization by resampling sequences of 1D tokens from pre-trained 2D VAE tokens. 
In contrast, our encoder builds on raw RGB images. More importantly, our approach is motivated by a fundamentally different objective --- reintroducing a PCA-like structure into visual tokenization to enforce a structured, hierarchical latent representation. 
Furthermore, our tokenizer is continuous rather than discrete, setting it apart from~\cite{bachmann2025flextok} and allowing it to better capture the variance-decaying properties inherent to PCA. 
Additionally, we identify and resolve the semantic-spectrum coupling effect, a key limitation in existing visual tokenization methods that have not been previously addressed.

\subsection{Related Work on Human Perception}

\textbf{Human perception} of visual stimuli has been shown to follow the global precedence effect~\cite{NAVON1977353}, where the global information of the scene is processed before the local information.
In~\cite{fei2007we}, controlled experiments of presentation time on human perception of visual scenes have further confirmed the global precedence effect, where less information (presentation time) is needed to access the non-semantic, sensory-related information of the scene compared to the semantically meaningful, object- or scene-related information.
Similar results have been reported in~\cite{inference}, where sensory attributes are more likely to be processed when the scene is blurred.
Moreover, \cite{OLIVA2000176} has suggested that reliable structural information can be quickly extracted based on coarse spatial scale information.
These results suggest that human perception of visual stimuli is hierarchical, where the global information of the scene is processed before the local information.
As we have shown in the main paper, \modelname can naturally emerge a similar hierarchical structure in the token sequence, where the first few tokens encode the global information of the scene and the following tokens encode the local information of the scene.
This hierarchical structure is provably PCA-like, similar to the hierarchical nature of human perception of visual stimuli.

\subsection{Related Work on Diffusion-Based Tokenizers}
The usage of a diffusion-based decoder has been explored by several works~\cite{evae,dito,ge2024divot}.
Zhao et al.~\cite{evae} proposed the usage of a diffusion-based decoder as a paradigm shift from single-step reconstruction of previous tokenizers to the diffusion-based iterative refinement process.
Chen et al.~\cite{dito} further scale this idea on more modern DiT~\cite{dit} architecture and describe the scaling law for such diffusion-based tokenizers.
Ge et al.~\cite{ge2024divot} applied this idea to a video tokenizer, enabling better reconstruction and understanding of video content.
However, these previous works overlook the benefit of the diffusion-based decoder in that it can disentangle the semantic content from the spectral information.
Additionally, these works still apply the 2D grid-based structure for encoding the image without considering the latent structure of the token space.

\section{Additional Implementation Details} \label{sec:impl_detail_ext}
\subsection{Semanticist Autoencoder}
\paragraph{Model architecture.}
As shown in \cref{fig:arch}, the \modelname tokenizer follows the diffusion autoencoder~\cite{diffae,RCG} paradigm: a visual encoder takes RGB images as input and encodes them into latent embeddings to condition a diffusion model for reconstruction. In our case, the visual encoder is a ViT-B/16~\cite{vit} with a sequence of concept tokens concatenated with image patches as input. The concept tokens have full attention with patch tokens, but are causal to each other. Before being fed to the decoder, the concept tokens also go through a linear projector, and are then normalized by their mean and variance. To stabilize training, we also apply drop path with a probability of 0.1 to the ViT.
For the DiT decoder, we concatenate the patch tokens (condition) with noisy patches as input, and the timesteps are still incorporated via AdaLN following common practice~\cite{dit}. 

\paragraph{Nested classifier-free guidance (CFG).}
For the DiT decoder, we randomly initialize $k$ (number of concept tokens) learnable null-conditioning tokens. During each training iteration, we uniformly sample a concept token index $k'$, and corresponding null tokens replace all tokens with larger indices. To facilitate the learning of the encoder, we do not enable nested CFG in the first 50 training epochs. During inference, CFG can be applied to concept tokens independently following the standard practice~\cite{dit}.

\paragraph{Training.}
We follow~\cite{RCG} for training details of the tokenizer. Specifically, the model is trained using the AdamW~\cite{adamw} optimizer on ImageNet~\cite{imagenet} for 400 epochs with a batch size of 2048. The base learning rate is 2.5e-5, which is scaled by $lr = lr_\text{base} \times \text{batch size}/256$. The learning rate is also warmed up linearly during the first 100 epochs, and then gradually decayed following the cosine schedule. No weight decay is applied, and $\beta_1$ and $\beta_2$ of AdamW are set to 0.9 and 0.95. During training, the image is resized so that the smaller side is of length 256, and then randomly flipped and cropped to 256$\times$256. We also apply a gradient clipping of 3.0 to stabilize training. The parameters of the model are maintained using exponential moving average (EMA) with a momentum of 0.999.

\paragraph{Inference.}
Because of the nature of the PCA structure, it is possible to obtain reasonable reconstruction results with only the first few concept tokens. In implementation, we achieve this by padding missing tokens with their corresponding null conditioning tokens and then feeding the full sequence to the DiT decoder.

\subsection{Autoregressive Modeling}

\paragraph{Model architecture.}
The \armodelname roughly follows the LlamaGen architecture with the only change of using a diffusion MLP as the prediction head instead of a softmax head.
To perform the classifier-free-guidance, we use one \texttt{[CLS]} token to guide the generation process of \armodelname.
As certain configurations of \modelname can yield high-dimensional tokens, we made a few adjustments to the model architecture of \armodelname to allow it to learn with high-dimensional tokens.
Specifically, we use a 12-layer MLP with each layer having 1536 hidden neurons as the prediction head and use the stochastic interpolant formulation~\cite{sit} to train the diffusion MLP.
The classifier-free guidance is also slightly modified: we concatenate the \texttt{[CLS]} token with the input to the diffusion MLP along the feature axis and then project back to the original feature dimension to feed into the diffusion MLP.
These changes allow us to train auto-regressive models on high-dimensional (e.g., 256-dimensional) tokens with improved stability compared to the original version proposed in~\cite{mar}.
However, we expect future research to drastically simplify this model architecture.

\paragraph{Training.}
The \armodelname is trained for 400 epochs with cached latents generated by pretrained \modelname on the ImageNet dataset with TenCrop and random horizontal flipping augmentations.
We use a batch size of 2048, and apply a 100-epoch warmup for the base learning rate of 1e-4, which is scaled similarly as the \modelname w.r.t. the batch size.
After warmup, the learning rate is fixed.
Weight decay of 0.05 and gradient clipping of 1.0 are applied.
In our experiments, we find that later concept tokens have diminishing returns or are even harmful for \armodelname, thus only train \armodelname with the first few tokens.
Specifically, the \armodelname-L model is trained with 32 concept tokens. %

\paragraph{Inference.}
In the inference stage, we use the same linear classifier-free guidance schedule as MAR~\cite{mar} and MUSE~\cite{muse}. The schedule tunes down the guidance scale of small-indexed tokens to improve the diversity of generated samples, thus being more friendly for gFID.
When reporting gFID, we disable CFG for \modelname's DiT decoder, tune the guidance scale of the autoregressive model, and report the best performance.

\subsection{Linear Probing}

We utilized the \texttt{sklearn} library to perform the linear probing experiments, the encoder weights are frozen, and we encode each image to its token representation. 
The linear classifier is learned on the token space without applying any data augmentation.

\section{Additional Experiment Results}

\subsection{Human Perception Test}
\label{sec:human_eval}

\begin{figure}
    \centering
    \includegraphics[width=.95\linewidth]{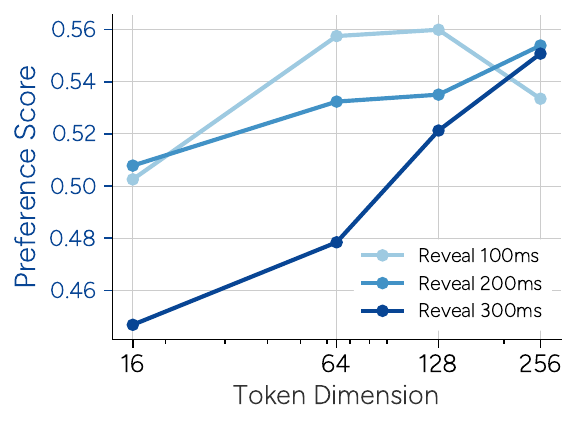}
    \caption{The preference score from the human perception test, all models and test configurations obtained a score close to 0.5, indicating \modelname can encode images as effectively as human language does.}
    \label{fig:humaneval_curve}
\end{figure}

We are interested in understanding whether the tokens learned by \modelname follow a human-like perception effect, namely the global precedence effect~\cite{NAVON1977353} where the global shape and semantics are picked up within a very short period of exposure.
Thus, we designed a human perception test to evaluate whether \modelname generates tokens that closely follow human perception.
Specifically, we generate images by only reconstructing from the first two tokens from \modelname.
Distractor images are also generated by first captioning the image with Qwen2.5VL~\cite{bai2025qwen2} and then generate the image with a stable-diffusion model~\cite{LDM}.
Following the setup of~\cite{fei2007we}, we only reveal the generated images and the distractors by a very short reveal time, and then ask the participants to choose which images more closely align with the original image.
For evaluation, we give the participant's preference to distractor image zero points, the preference to the generated image one point, and in the case of a tie, we give 0.5 points.
\cref{fig:humaneval_curve} presents the averaged preference score with different token dimensions and reveal time.
\modelname is able to obtain a score close to 0.5 under all cases, indicating that \modelname can encode the image's global semantic content close to how state-of-the-art vision language models~\cite{bai2025qwen2} encode the image in language space.
A web-based human perception test interface is provided along with this appendix.

\subsection{Zero-Shot CLIP on Reconstructed Images}
\begin{figure}
    \centering
    \includegraphics[width=.9\linewidth]{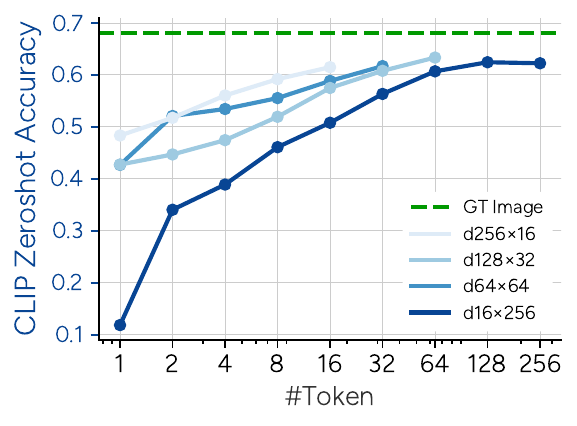}
    \caption{CLIP zero-shot accuracy on reconstructed images.}
    \label{fig:clip_zs}
\end{figure}

We also study the property of the \modelname latent space by reconstructing from it.
\cref{fig:clip_zs} demonstrate the zero-shot accuracy of a pretrained CLIP~\cite{radford2021clip} model on the imagenet validation set reconstructed by \modelname.
For all model variants, the zero-shot performance improves with the number of tokens, with models using more dimensions per token achieving better performance with a smaller number of tokens, indicating that with more dimensions, \modelname is able to learn the semantic content with fewer tokens.
\cref{fig:dit_scale} provides the rFID score on the ImageNet validation set with a varying number of tokens, similar conclusions can be drawn.
Additionally, \cref{fig:dit_scale} also provides the scaling behavior of \modelname, we can observe that \modelname not only enjoys a structured latent space, but also demonstrates a promising scaling.

\subsection{Semantic Spectrum Coupling Effect Results}
\begin{figure}[t]
    \centering
    \includegraphics[width=.9\linewidth]{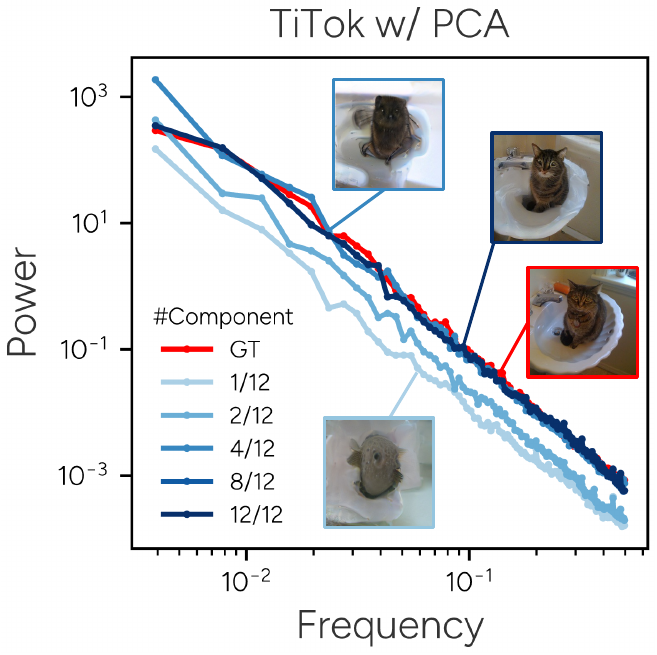}
    \caption{Frequency-power spectra of TiTok decomposed with PCA at feature dimensions. The learning of semantic contents and spectral information is coupled.}
    \label{fig:spectral_titok_pca}
\end{figure}

In \cref{fig:spectral_titok_pca}, we present the power frequency plot of performing PCA to decompose the latent token space of TiTok~\cite{titok}.
A similar effect as the PCA decomposition on VQ-VAE~\cite{llamagen} and the first $k$ token decomposition on TiTok~\cite{titok} is observed.
This result further demonstrates that the latent space of TiTok~\cite{titok} entangles the semantic contents and the spectral information.

\begin{figure*}[t]
    \centering
    \includegraphics[width=\linewidth]{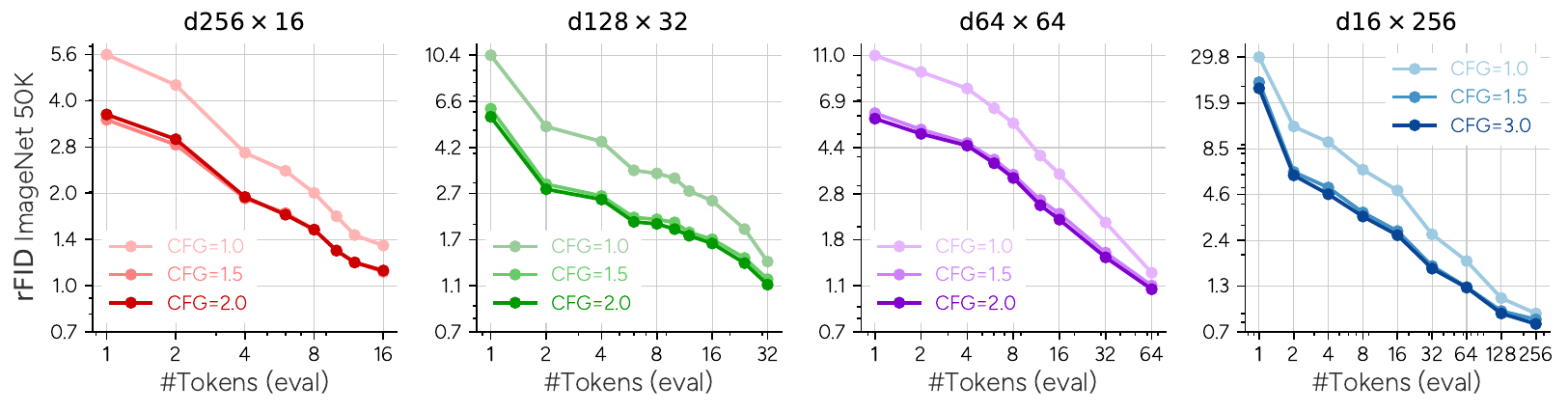}
    \caption{Reconstruction performance of different encoder configurations on ImageNet val 50K benchmark. A larger number of lower-dimensional tokens is more friendly for reconstruction tasks.} \label{fig:rfid_imagenet50k_dims}
\end{figure*}

\begin{figure}[t]
    \centering
    \includegraphics[width=.9\linewidth]{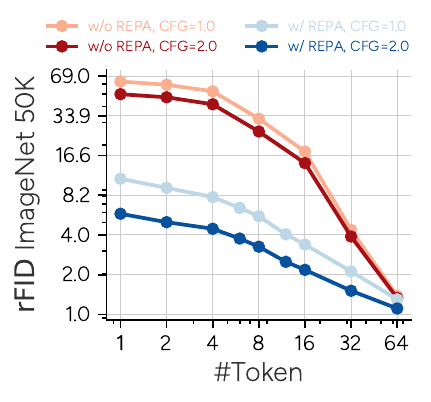}
    \caption{Ablation on the use of REPA (with d64$\times$64 concept tokens, DiT-L/2 decoder, see qualitative results in \cref{fig:recon_64dim_repa}). REPA improves the information density in preceding tokens.}
    \label{fig:repa_ablation}
\end{figure}

\subsection{Additional Ablation Study}\label{sec:ablation_ext}

In \cref{fig:repa_ablation}, we show the results of \modelname with d64$\times$64 tokens trained with or without REPA~\cite{repa} evaluated by reconstruction FID on ImageNet 50K validation set. Despite the performance with full tokens being similar, adding REPA significantly improves the contribution of each (especially the first few) tokens. This naturally fits our need for PCA-like structure and is thus adopted as the default.

We also compared the reconstruction performance of different concept token dimensions. We fix the product between the number of tokens and the dimension per token to be 4096, and investigate 256-dimensional (d256$\times$16), 128-dimensional (d128$\times$32), 64-dimensional (d64$\times$64), and 16-dimensional (d16$\times$256) tokens. As shown in~\cref{fig:rfid_imagenet50k_dims}, all configurations can learn ordered representations, with higher-dimensional ones containing more information per token. However, lower-dimensional tokens are more friendly for reconstruction tasks as they achieve better rFID.

\begin{figure*}[t]
    \centering
    \includegraphics[width=\linewidth]{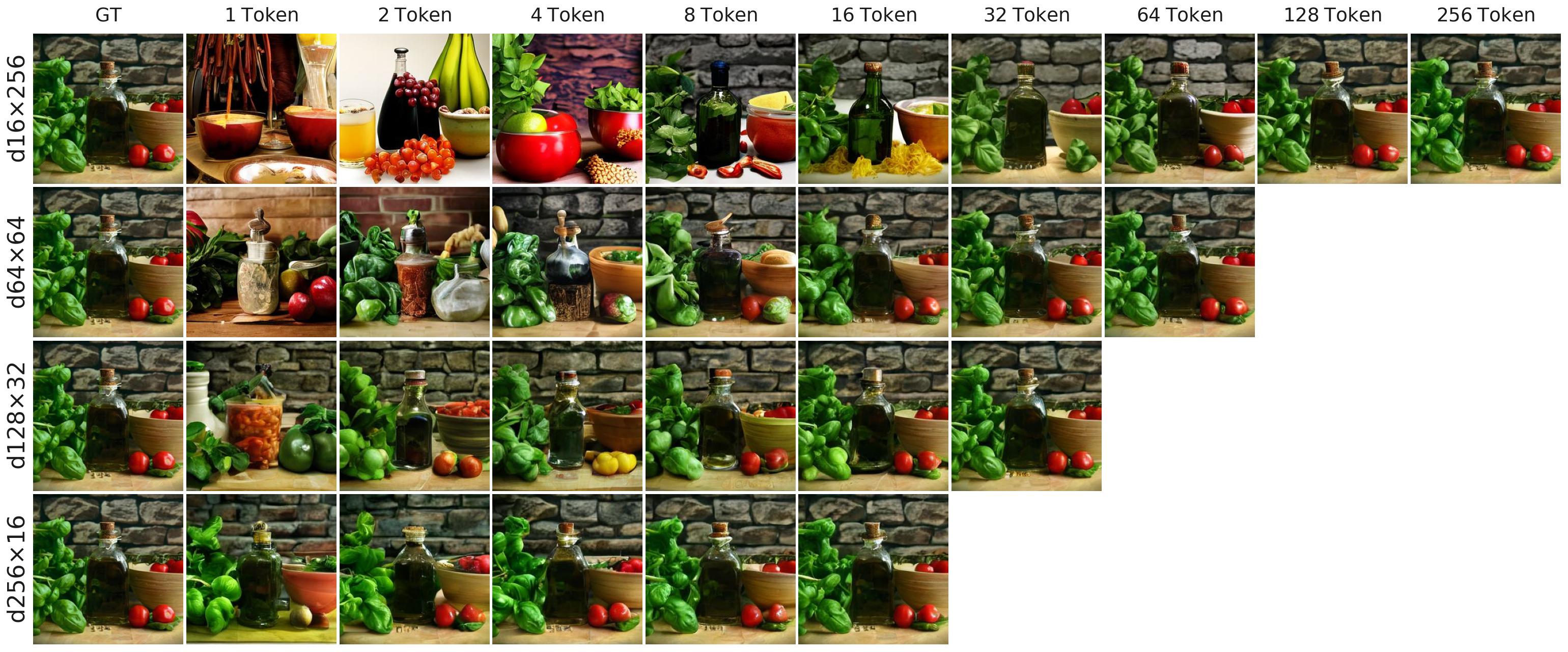}
    \caption{Qualitative results of different token dimensions. Higher-dimensional tokens encode more information, and lower-dimensional tokens achieve clearer semantic decoupling and better reconstruction.}
    \label{fig:recon_dim_comparison}
\end{figure*}

\begin{figure*}[t]
    \centering
    \includegraphics[width=\linewidth]{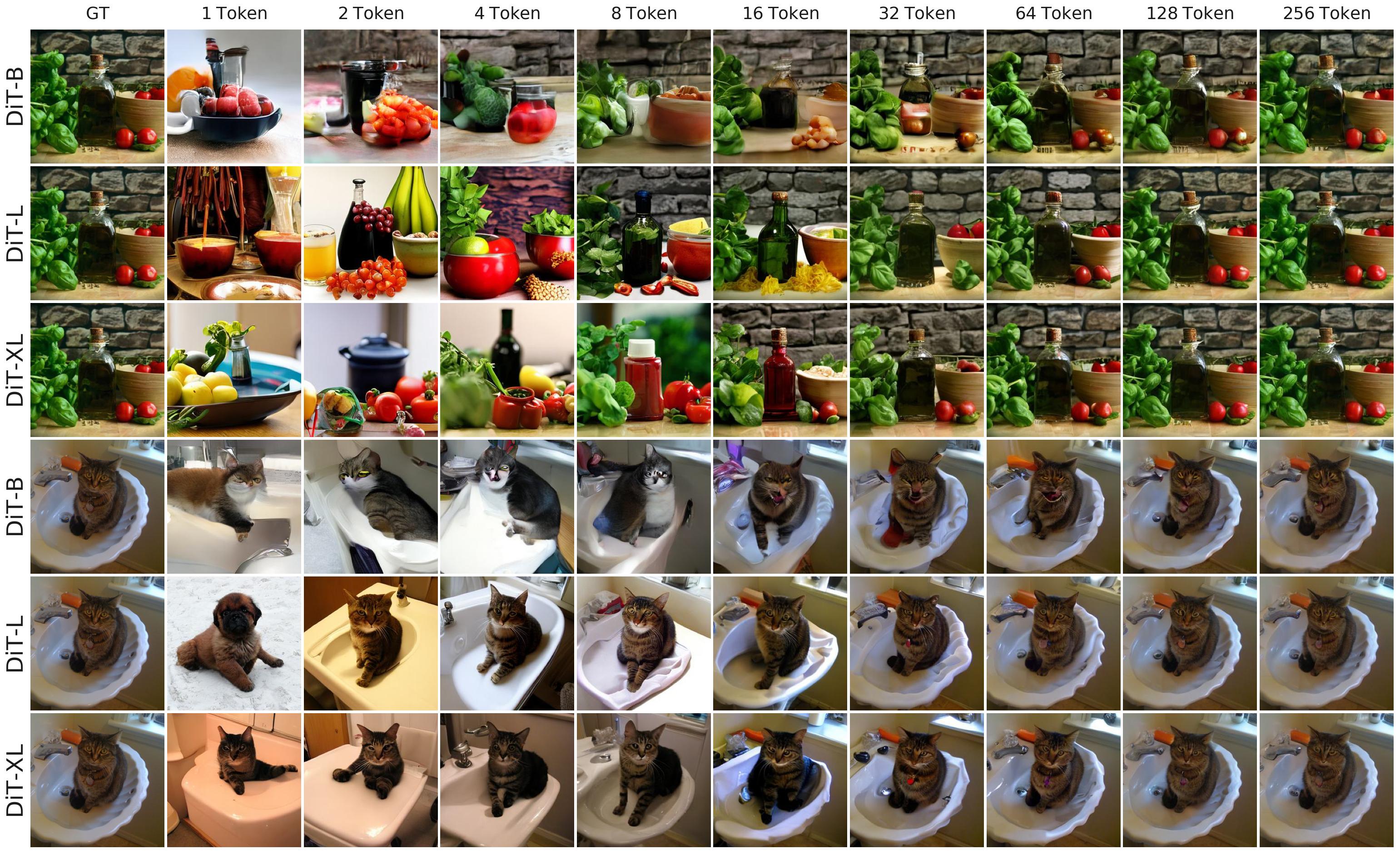}
    \caption{Qualitative results of different DiT decoder scales (DiT-B/2, DiT-L/2, and DiT-XL/2) with d16$\times$256 tokens. The quality of images generated with fewer tokens improves consistently as the decoder scales up.}
    \label{fig:recon_16dim_scale}
\end{figure*}

\begin{figure*}[t]
    \centering
    \includegraphics[width=\linewidth]{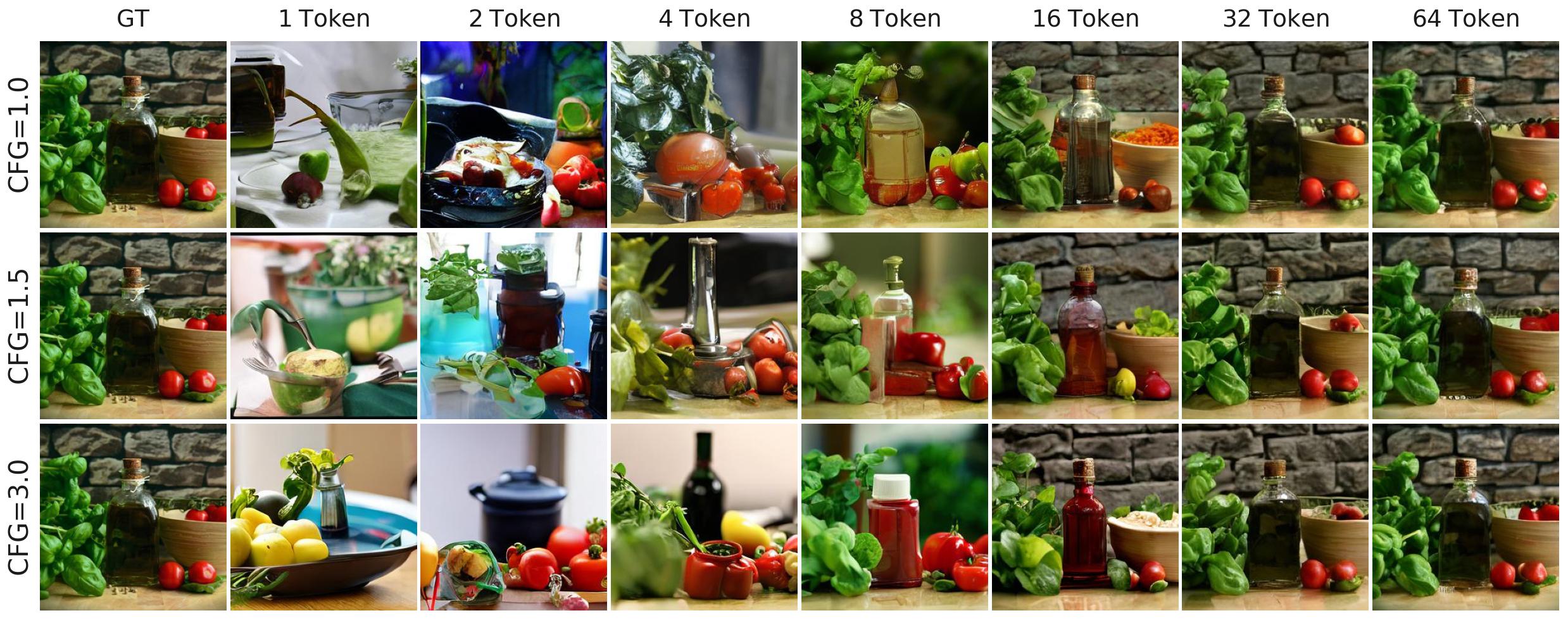}
    \caption{Qualitative results of different CFG guidance scales for DiT decoder, which clearly controls image aesthetics.}
    \label{fig:recon_16dim_cfg}
\end{figure*}

\subsection{Qualitative Results}

In~\cref{fig:recon_dim_comparison}, reconstruction results from using different numbers of token dimensions are presented.
As the dimension for one token becomes large, more semantic content can be encoded into it, thus allowing \modelname to generate faithful reconstructions of the original image.

In~\cref{fig:recon_16dim_scale}, the reconstructed results for different scaled DiT decoders are presented. 
These models are trained with the same dimension for the tokens that are 16-dimensional.
We can see that as the model scales up, the reconstructed images with fewer tokens become more and more realistic and appealing.

\cref{fig:recon_16dim_cfg} shows the reconstruction of the same \modelname tokenizer with different CFG guidance scales at inference time (CFG=1.0 indicates not applying CFG). It can be seen that the guidance scale has a very strong correlation with the aesthetics of generated images.

\begin{figure*}[t]
    \centering
    \includegraphics[width=\linewidth]{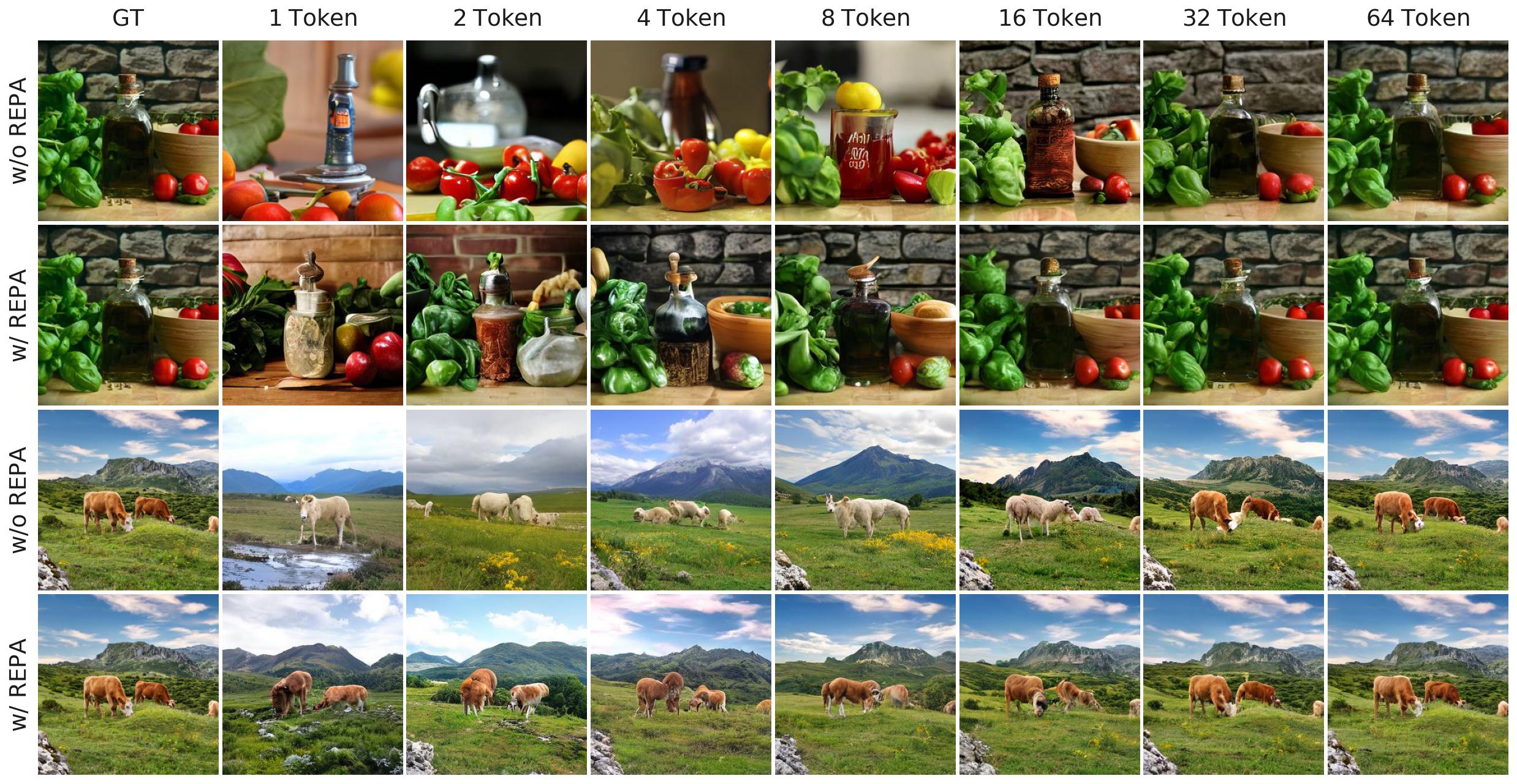}
    \caption{Qualitative results on effects of REPA (with d64$\times$64 concept tokens). Instead of improving final reconstruction much, the benefit of REPA is mainly attributed to more faithful semantics in intermediate results.}
    \label{fig:recon_64dim_repa}
\end{figure*}

\cref{fig:recon_64dim_repa} presents qualitative results with or without the usage of REPA~\cite{repa}.
It is clear that the usage of REPA did not visually improve the final reconstruction by much, yet with fewer tokens, the model with REPA demonstrates more faithful semantic details with the original image.

\begin{figure*}[t]
    \centering
    \includegraphics[width=\linewidth]{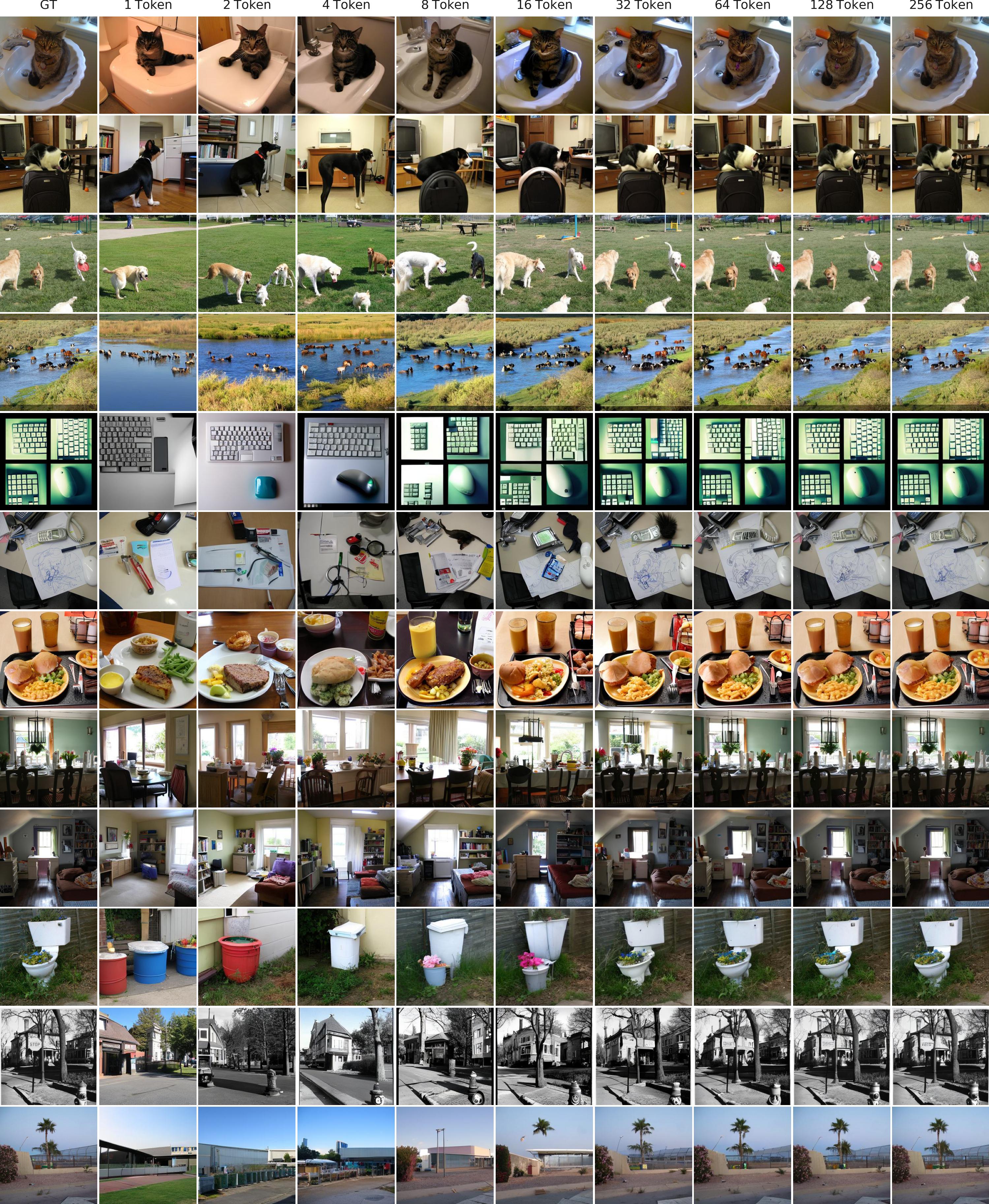}
    \caption{More reconstruction results of \modelname autoencoder (with d16$\times$256 concepts tokens and DiT-XL/2 decoder).}
    \label{fig:teaser_extend}
\end{figure*}

\begin{figure*}
    \centering
    \includegraphics[width=0.7\linewidth]{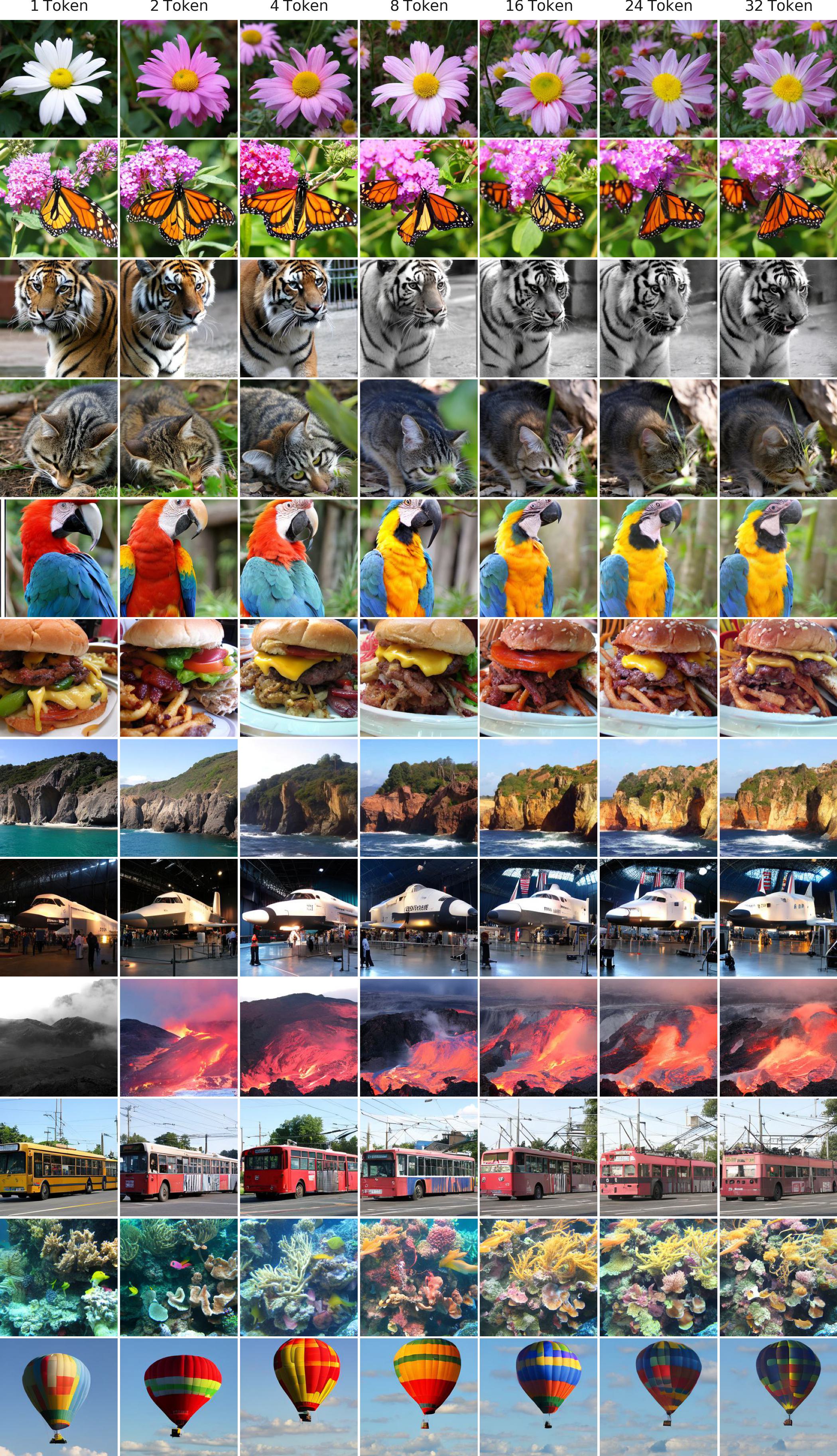}
    \caption{More visualization of intermediate results of auto-regressive image generation.}
    \label{fig:more_example_ar}
\end{figure*}

\cref{fig:teaser_extend} demonstrates the reconstruction results of more randomly sampled images, and \cref{fig:more_example_ar} illustrates more intermediate results of auto-regressive image generation.

